\documentclass{article}
\PassOptionsToPackage{numbers, compress}{natbib}
\usepackage[main, final]{neurips_2026}

\usepackage[utf8]{inputenc}
\usepackage[T1]{fontenc}
\usepackage{hyperref}
\usepackage{url}
\usepackage{booktabs}
\usepackage{amsfonts}
\usepackage{amsmath}
\usepackage{nicefrac}
\usepackage{microtype}
\usepackage{graphicx}
\usepackage{subcaption}
\usepackage[table]{xcolor}
\usepackage{array}
\usepackage{amssymb}
\newcommand{\cmark}{\ensuremath{\checkmark}}
\newcommand{\xmark}{\ensuremath{\times}}
\usepackage{multirow}
\usepackage{algorithm}
\usepackage{algpseudocode}
\usepackage{float}
\usepackage{siunitx}
\usepackage{textcomp}
\usepackage{makecell}
\usepackage{tabularx}
\usepackage{longtable}
\usepackage{colortbl} 

\newcommand{\projectlink}{\texttt{https://xiaoang0211.github.io/ProMo3D/}}

\title{ProMo3D: Probing Motion Cues in Frozen 3D Foundation Models}

\author{%
Xiaoang Zhang$^{1}$\thanks{This work was conducted while Xiaoang Zhang was a guest researcher at the Technical University of Munich.}
\quad Mert Kiray$^{1,2,3}$
\quad Yordanka Velikova$^{1,2}$\\
\textbf{Manoj Biswanath$^{1,2}$ \quad Benjamin Busam$^{1,2,3}$}\\[6pt]
$^{1}$Technical University of Munich (TUM), Germany\\
$^{2}$Munich Center for Machine Learning (MCML), Germany\\
$^{3}$Obsphera, Munich, Germany
}

\begin{document}

\maketitle
\begin{abstract}
Feed-forward 3D foundation models have recently reshaped multi-view reconstruction, yet dynamic 4D understanding still often relies on large-scale motion supervision or hand-crafted, architecture-specific heuristics. We ask whether frozen 3D foundation model tokens already encode motion-relevant information that can be exposed by lightweight readout functions, rather than re-learning motion-specific features with high-capacity decoders. Through probing on three representative backbones, we find that static--dynamic separation is decodable from frozen token representations. Our rank-restricted and spectral analyses on $\pi^3$X reveal that the recovered motion cues are concentrated in a low-dimensional subspace. Building on this, we propose Probing-based Motion Distillation (PMD), which converts pairwise motion-state compatibility scores against a fixed anchor token into dense motion masks. Trained on only 160 YouTube-VOS videos using a single 16 GB GPU, PMD transfers zero-shot to DAVIS, SegTrackv2, and FBMS-59 without target-benchmark fine-tuning. With multi-view pretrained geometric backbones, PMD outperforms training-free heuristics and approaches substantially heavier supervised and self-supervised methods, suggesting that motion decoding from frozen 3D foundation models can be lightweight and data-efficient.
\par\smallskip\noindent\textbf{Project page:} \projectlink
\end{abstract}

\section{Introduction}

\begin{figure}[!htbp]
    \centering
    \includegraphics[width=\linewidth]{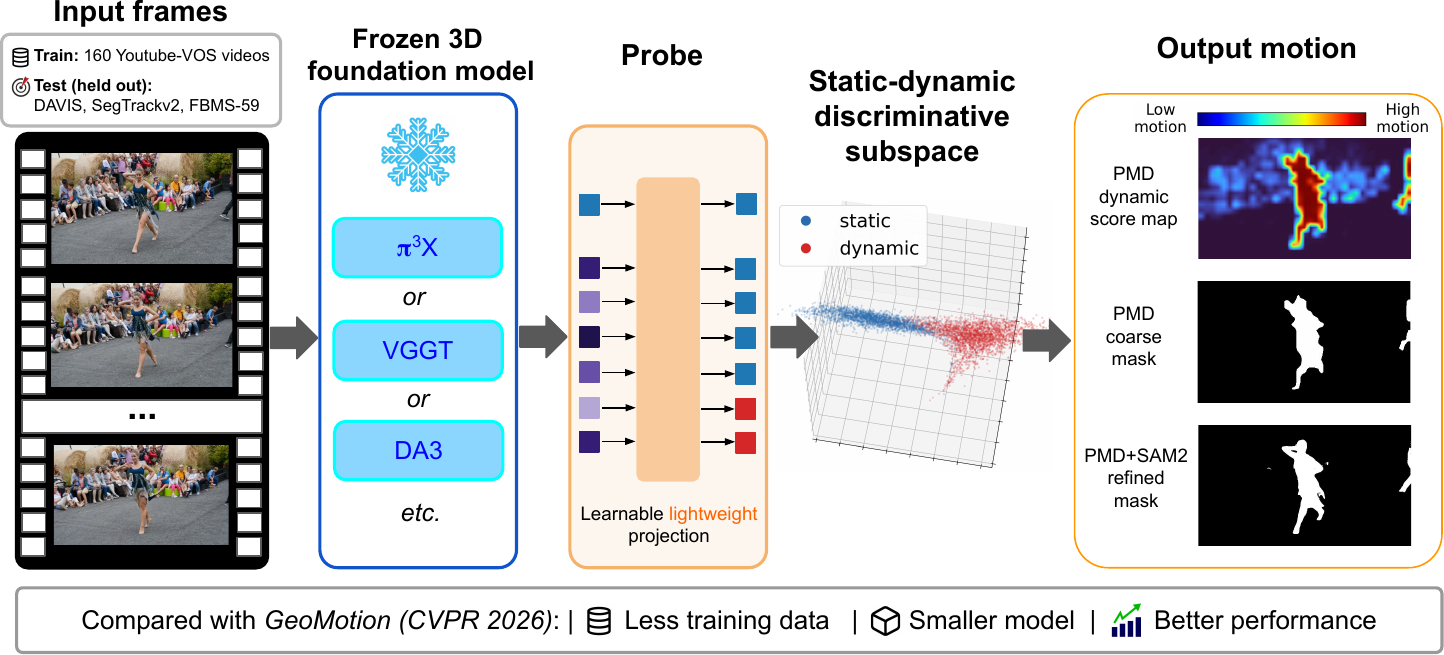}
    \caption{
    Overview of Probing-based Motion Distillation (PMD). Given an input frame group, a frozen feed-forward 3D foundation model such as $\pi^3$X~\citep{pi3}, VGGT~\citep{vggt}, or Depth Anything 3 (DA3)~\citep{da3} produces token representations. A lightweight probe trained on only 160 videos maps these features into a probe-induced static--dynamic discriminative representation. PMD then converts anchor-based pairwise probe scores into dense dynamic score maps, which are thresholded into coarse masks and optionally refined with SAM2~\citep{sam2}. It transfers zero-shot to DAVIS~\citep{davis16,davis17}, SegTrackv2~\citep{segtrack}, and FBMS-59~\citep{fbms}, while requiring substantially less training data and trainable capacity than high-capacity motion decoders such as GeoMotion~\citep{geomotion}.
    }
    \label{fig:teaser_qualitative}
\end{figure}

Driven by large-scale geometry pretraining, feed-forward 3D foundation models have recently reshaped 3D vision. Methods such as DUSt3R~\citep{dust3r}, VGGT~\citep{vggt}, $\pi^3$/$\pi^3$X~\citep{pi3}, and Depth Anything 3~\citep{da3} bypass traditional iterative optimization and recover 3D geometry directly from unposed multi-view images.

Despite these advances, dynamic 4D understanding remains challenging because static geometry, camera-induced motion, and independently moving objects must be disentangled. Existing approaches typically follow two paradigms. Large-scale supervised methods~\citep{d4rt,ufo4d,page4d} learn motion and object tracking from dense annotations but require substantial labeling effort and high computational costs. Training-free alternatives~\citep{vggt4d,easi3r} extract motion cues from frozen 3D foundation models using hand-crafted, architecture-specific heuristics such as attention layers or key--query Gram matrices. While efficient, these heuristics are often brittle and underperform learning-based motion decoders by a large margin, as shown by \citet{geomotion}.

Several observations suggest a third paradigm: exploiting motion cues already encoded in frozen 3D models to bridge the gap between efficiency and performance. First, the increasing use of 4D training data from multi-domain resources~\citep{waymo_open,kubric,omniworld} suggests that large feed-forward 3D backbones may acquire motion-relevant cues during geometry pretraining. Second, their implicit multi-view geometric representations appear to remain robust in dynamic scenes: recent work~\citep{d4rt,vggt4d,page4d} shows competitive camera pose estimation under scene motion, while \citet{geomotion} demonstrates that latent 4D geometry from $\pi^3$~\citep{pi3} provides a strong prior for motion segmentation, albeit with a high-capacity neural network-based decoder. In addition, \citet{bratulic} shows that intermediate layers of VGGT encode geometric constraints such as correspondences and epipolar geometry. This suggests that regions violating multi-view geometric consistency may also be detectable in frozen token representations. Taken together, these findings motivate our central inquiry: can motion cues be recovered from frozen 3D foundation model tokens using low-capacity probes, without complex decoders or expensive fine-tuning?

We investigate this question through supervised representation probing~\citep{alain,belinkov}. By training lightweight probes on frozen 3D foundation models, we find that motion cues within token representations can be mapped into a compact, probe-induced static--dynamic discriminative subspace that generalizes across diverse dynamic scenarios. Building on this finding, we introduce Probing-based Motion Distillation (PMD), which converts pairwise motion-state compatibility scores into dense motion masks.

Our contributions are summarized as follows:
\begin{itemize}
    \item \textbf{Systematic localization of motion cues in 3D backbones.} We conduct an extensive empirical study on three representative feed-forward 3D foundation models: $\pi^3$X~\citep{pi3}, Depth Anything 3~\citep{da3}, and DUSt3R~\citep{dust3r}. We identify the layers in which motion cues emerge in these frozen backbones, showing that they can be exposed by probing without relying on architecture-specific heuristics.
    
   \item \textbf{Characterization of a low-rank motion subspace.} Through rank-restricted probing and spectral analysis on $\pi^3$X, we find the probe-induced motion representation to be remarkably compact. We show that a low-rank quadratic probe is sufficient to extract dense motion masks, suggesting that motion understanding in 3D foundation models is largely a lightweight decoding problem rather than one requiring extensive feature re-learning.

   \item \textbf{Data-efficient Probing-based Motion Distillation (PMD).}
    We introduce PMD, a lightweight framework trained on 160 YouTube-VOS~\citep{youtubevos} videos using a single consumer-grade GPU with 16\,GB VRAM. PMD transfers zero-shot to standard motion segmentation benchmarks without target-benchmark fine-tuning. With multi-view pretrained geometric backbones, it outperforms training-free heuristics and remains competitive with supervised and self-supervised methods, while requiring substantially less training data.
\end{itemize}

\section{Related Work}

\subsection{Feed-forward 3D Foundation Models}

Feed-forward 3D foundation models such as DUSt3R~\citep{dust3r}, VGGT~\citep{vggt}, and $\pi^3$~\citep{pi3} have reshaped multi-view reconstruction by directly predicting 3D geometry and camera parameters from unposed images, bypassing iterative Structure-from-Motion pipelines~\citep{sfm}. Extensions such as Dens3R~\citep{dens3r}, AnySplat~\citep{anysplat}, and Depth Anything 3~\citep{da3} broaden this paradigm toward dense geometry, feed-forward 3D Gaussian reconstruction, and ray prediction. As recent backbones are increasingly trained on diverse geometric data, including dynamic sources~\citep{kubric,waymo_open,omniworld}, they provide a potential substrate for extracting motion cues.

\subsection{Dynamic Scene Reconstruction and Motion Segmentation}

Dynamic scene reconstruction remains an active research area due to challenges in disentangling geometry, camera motion, and object motion in diverse scenarios. NeRF-based approaches~\citep{dnerf,nsff} optimize scene-specific dynamic representations, often with auxiliary priors such as optical flow, depth, or masks. Recent feed-forward 4D methods~\citep{d4rt,ufo4d,page4d} move toward direct dynamic-scene reconstruction by predicting geometry, camera pose, and motion-related quantities from unposed views. A parallel line of work~\citep{em_ofs,oclr,amc_net,ocmos} focuses on motion segmentation, separating independently moving regions from static scene content using motion--appearance cues, optical flow, object-centric grouping, or appearance refinement.

Recent methods also attempt to extract motion cues from frozen 3D foundation models. Training-free approaches such as Easi3R~\citep{easi3r} and VGGT4D~\citep{vggt4d} rely on hand-crafted heuristics, including attention patterns or key--query Gram matrices. While computationally efficient, these heuristics are typically brittle, architecture-specific, and show a substantial performance gap compared to learned motion segmentation approaches, as shown by \citet{geomotion}. Most closely related to our motivation, \citet{geomotion} show that latent 4D geometry from $\pi^3$ provides a strong prior for motion segmentation. However, their approach still relies on a high-capacity neural decoder and additional optical-flow cues. Our work targets the gap between these two regimes: training-free heuristics are often brittle, whereas learned decoders are substantially more complex and expensive to train.

\subsection{Representation Probing and Latent Analysis}

Representation probing~\citep{alain,belinkov} tests whether specific information is decodable from neural representations. Recent studies show that pretrained vision and world models can contain structured latent information not explicit in the training objective: \citet{leworldmodel} probe physical quantities from JEPA-style latent representations, and \citet{bratulic} show that VGGT encodes geometric priors such as correspondences and epipolar constraints. Most closely related, \citet{objectbinding} show that object binding emerges in self-supervised ViT encoders and can be most effectively decoded using quadratic probes. We adopt this probing perspective for motion understanding, using low-capacity probes to test whether frozen 3D representations can be mapped into a static--dynamic discriminative subspace.

\section{Method}
Given an input frame group, a frozen feed-forward 3D foundation model produces target-frame token features at block \(\ell\):
\begin{equation}
\mathbf{H}^{(\ell)}=\{\mathbf{h}_0^{(\ell)}, \mathbf{h}_1^{(\ell)}, \dots, \mathbf{h}_N^{(\ell)}\},
\qquad \mathbf{h}_i^{(\ell)} \in \mathbb{R}^d .
\end{equation}
Here, \(N\) is the number of target-frame tokens and \(d\) is the feature dimension. Adjacent frames provide temporal or multi-view context, while token pairs are sampled only within the target frame.

Each token is assigned a binary motion label \(y_i\in\{0,1\}\) by projecting the dynamic mask onto the token grid, where \(y_i=1\) denotes a dynamic token and \(y_i=0\) denotes a static token. Following \citet{objectbinding}, we formulate motion probing as pairwise classification. For a token pair \((i,j)\), we define
\begin{equation}
z_{ij}=\mathbb{I}[y_i=y_j],
\end{equation}
where \(\mathbb{I}[\cdot]\) is the indicator function, which equals 1 if the condition is true and 0 otherwise. Thus, \(z_{ij}=1\) indicates that two tokens share the same motion state, and \(z_{ij}=0\) indicates different motion states.

A probe predicts the pairwise probability \(\hat z_{ij}\in[0,1]\). All probes are trained on sampled token pairs \(\mathcal{P}\) using binary cross-entropy while keeping the backbone parameters frozen:
\begin{equation}
\mathcal{L}_{\mathrm{probe}}
= -\frac{1}{|\mathcal{P}|}
\sum_{(i,j)\in\mathcal{P}}
\left[ z_{ij}\log \hat z_{ij} + (1-z_{ij})\log(1-\hat z_{ij}) \right].
\end{equation}

\subsection{Pairwise Probe Functions}

Let \(\mathbf{h}_i,\mathbf{h}_j\in\mathbb{R}^d\) denote two token features. We evaluate three lightweight probe functions that differ in how they model pairwise motion-state compatibility: independent linear terms, channel-wise multiplicative interactions, and cross-channel bilinear interactions.

\paragraph{Linear Probe.}
The linear probe tests whether motion-state compatibility is directly decodable from individual token features:
\begin{equation}
\hat z_{ij}^{\mathrm{lin}}
=
\sigma\!\left(
\mathbf{w}_1^\top \mathbf{h}_i
+
\mathbf{w}_2^\top \mathbf{h}_j
+
b
\right),
\end{equation}
where \(\mathbf{w}_1,\mathbf{w}_2\in\mathbb{R}^d\) and \(b\in\mathbb{R}\) are learnable parameters.

\paragraph{Diagonal Quadratic Probe.}
The diagonal quadratic probe captures channel-wise multiplicative interactions:
\begin{equation}
\hat z_{ij}^{\mathrm{diag}}
=
\sigma\!\left(
\mathbf{h}_i^\top \mathbf{W}_{\mathrm{diag}}\mathbf{h}_j
+
b
\right),
\qquad
\mathbf{W}_{\mathrm{diag}}
=
\frac{1}{\sqrt d}\operatorname{diag}(\boldsymbol{\alpha}),
\end{equation}
where \(\boldsymbol{\alpha}\in\mathbb{R}^d\).

\paragraph{Low-Rank Quadratic Probe.}
Our primary probe is a low-rank bilinear form:
\begin{equation}
\label{eq:low_rank}
\hat z_{ij}^{\mathrm{lr}}
=
\sigma\!\left(
\mathbf{h}_i^\top \mathbf{W}_{\mathrm{lr}}\mathbf{h}_j
+
b
\right),
\qquad
\mathbf{W}_{\mathrm{lr}}
=
\frac{\mathbf{U}^\top\mathbf{V}+\mathbf{V}^\top\mathbf{U}}{2\sqrt d},
\end{equation}
where \(\mathbf{U},\mathbf{V}\in\mathbb{R}^{k\times d}\) and \(k\ll d\). Here, \(k\) is the factor-rank hyperparameter; symmetrization gives \(\operatorname{rank}(\mathbf W_{\mathrm{lr}})\leq 2k\). The symmetrized form matches the symmetric pair label \(z_{ij}=z_{ji}\) and enforces order invariance. By varying the rank \(k\), we test whether motion cues can be decoded through a compact, probe-induced low-rank bilinear representation.

\subsection{Probing-based Motion Distillation}

\emph{Probing-based Motion Distillation} (PMD) converts the learned pairwise compatibility into a dense motion estimator. While the probe is trained to predict the motion-state compatibility between arbitrary token pairs, PMD additionally formulates a unary approach by designating an anchor token $\mathbf{h}_a$ with a fixed spatial index $a$ across all frames.

Specifically, we use the first token at index $a=0$ as the fixed anchor token $\mathbf{h}_a$ throughout all experiments. For padded backbones~\citep{vggt,pi3}, this token lies in the padded region and is therefore static by construction; this covers the strongest PMD variants in our experiments (see Table~\ref{tab:sota_full}). For non-padded backbones~\citep{dust3r,da3,mapanything}, we keep the same fixed-index convention and analyze this choice empirically in Appendix~\ref{app:convention}. Querying the trained probe $f(\mathbf{h}_i, \mathbf{h}_j) \triangleq \hat z_{ij} \in [0,1]$ against this anchor yields a token-level static score:
\begin{equation}
\hat p_i^{\mathrm{stat}} = f(\mathbf{h}_a, \mathbf{h}_i), \qquad a=0.
\end{equation}
The dynamic score is then defined as
\begin{equation}
\hat p_i^{\mathrm{dyn}} = 1 - \hat p_i^{\mathrm{stat}} .
\end{equation}

The resulting token-level dynamic scores are reshaped to the token grid and bilinearly upsampled to the original image resolution. Since all anchor--target scores can be evaluated in parallel, PMD adds only lightweight computation on top of frozen backbone inference.

\section{Experiments}

\subsection{Experimental Setup}
\label{sec:setup}

All experiments, including backbone feature extraction, probe training, and evaluation, are conducted on a single consumer-grade NVIDIA RTX 5070 Ti GPU with 16GB VRAM.

\paragraph{Training Protocol.}
 Unless otherwise specified, all probes are trained on 160 videos from the official YouTube-VOS 2018 training split~\citep{youtubevos} and validated on 80 videos from its validation split. No data from the evaluation benchmarks (DAVIS~\citep{davis16,davis17}, SegTrackv2~\citep{segtrack}, or FBMS-59~\citep{fbms}) is used for training and validation.

\paragraph{Datasets and Metrics.}
We evaluate one binary mask per frame on DAVIS~\citep{davis16,davis17}, SegTrackv2~\citep{segtrack}, and FBMS-59~\citep{fbms}, reporting Jaccard mean (J-M) and recall (J-R), the fraction of evaluated frames with Jaccard score above 0.5. DAVIS16 maps valid foreground to dynamic and background to static. DAVIS17/DAVIS-all union all valid object IDs as dynamic, retain valid background as static, and exclude void pixels. SegTrackv2 unions separate foreground instance masks as dynamic and labels background as static. FBMS-59 labels valid pixels differing from the sequence-specific background label as dynamic and evaluates only annotated frames. These annotation-defined targets need not coincide with instantaneous physical motion. Token-wise score maps are bilinearly upsampled and binarized to obtain coarse PMD masks, with optional SAM2~\citep{sam2} refinement. For token-level probing analysis, we additionally report dynamic-token accuracy ($\mathrm{Acc}_{\mathrm{dyn}}$) and Normalized Gain (NG):
\[
\mathrm{Acc}_{\mathrm{dyn}}
=
\frac{1}{N}
\sum_{i=1}^{N}
\mathbf{1}\!\left[
\mathbf{1}(p_i \ge 0.5)=\mathbf{1}(r_i \ge \tau)
\right],
\qquad
\mathrm{NG}
=
\frac{\mathrm{Acc}_{\mathrm{dyn}}-\mathrm{Baseline}}{1-\mathrm{Baseline}} .
\]
Here, $N$ denotes the total number of evaluated tokens, $p_i$ is the predicted dynamic score, and $r_i \in [0,1]$ is the ground-truth dynamic pixel ratio within the image patch of token $i$. We use a threshold of $\tau=0.35$ to label a token as dynamic. Because static regions dominate our evaluation benchmarks, the baseline predicts all tokens as static; detailed class-imbalance statistics are provided in Appendix~\ref{app:imbalance}.

The selectivity control reports pairwise compatibility accuracy and balanced accuracy. Cross-backbone source-validation analysis reports pairwise normalized gain ($\mathrm{NG}_{\mathrm{pair}}$) against the always-same-state baseline (Appendix~\ref{app:backbone_disparity}). Implementation and thresholding details are in Appendix~\ref{app:implementation}.

\subsection{Layer-wise Motion Cue Emergence}
\label{sec:origin}
\begin{figure}[!htbp]
    \centering
    \includegraphics[width=\linewidth]{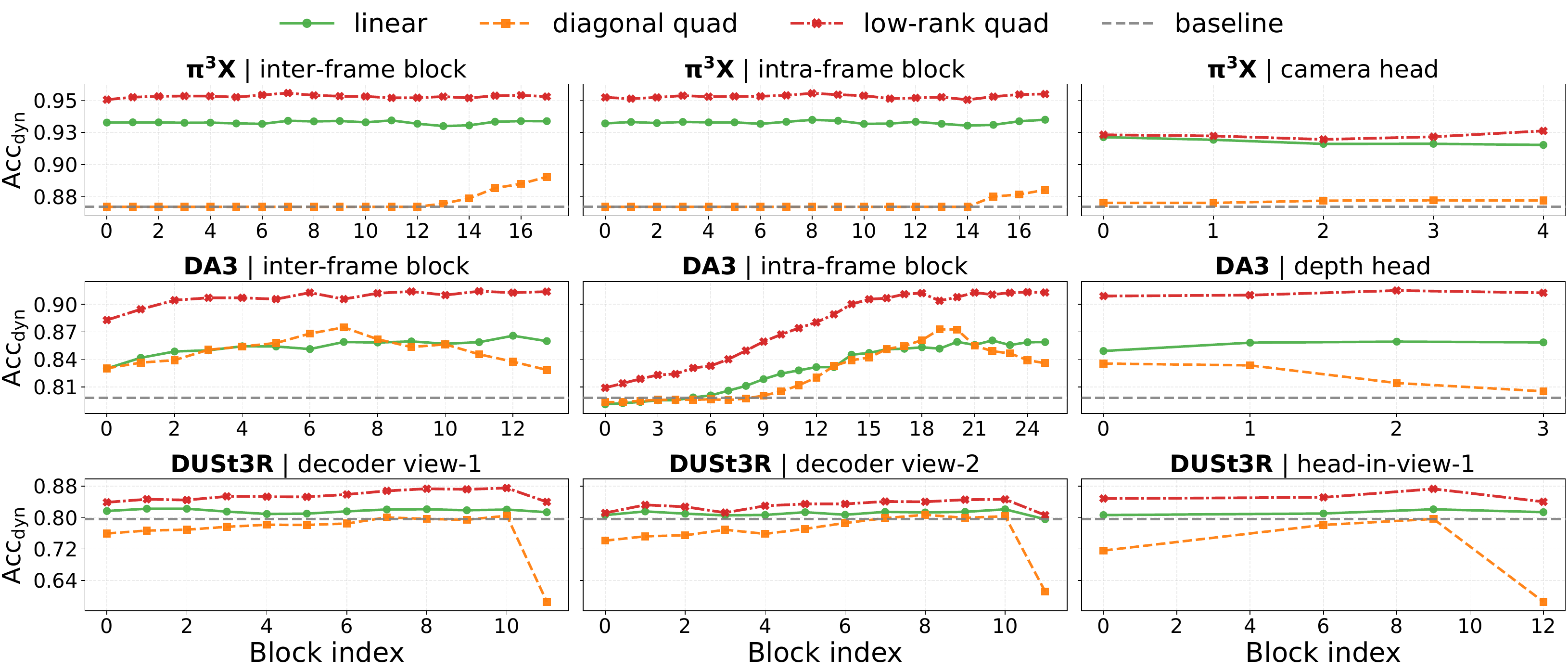}
    \caption{Layer-wise motion decodability on DAVIS-all~\citep{davis17}. Low-rank quadratic probes expose static--dynamic discriminative cues across the multi-view geometric modules of all three backbones, outperforming linear and diagonal probes.}
    \label{fig:layer_emergence}
\end{figure}

We perform layer-wise probing on three representative backbones: $\pi^3$X~\citep{pi3}, Depth Anything 3 (DA3)~\citep{da3}, and DUSt3R~\citep{dust3r}. Our probes operate on full block-output token representations. For $\pi^3$X and DA3, the selected block groups follow the VGGT-style alternating-attention design, consisting of frame-wise self-attention and global self-attention~\citep{vggt}. We refer to the corresponding frozen block-output token features as \emph{intra-frame} and \emph{inter-frame} features. As shown in Figure~\ref{fig:layer_emergence}, motion decodability is widespread across the multi-view geometric modules. The low-rank quadratic probe consistently outperforms the linear and diagonal probes. Motion cues also propagate to intra-frame blocks and prediction heads, but their quality varies across backbones. For subsequent experiments, PMD averages all blocks in one fixed backbone-specific group, using the settings in Table~\ref{tab:pmd_default} for every target sequence and dataset.

\begin{table}[!htbp]
\centering
\setlength{\tabcolsep}{3pt}
\caption{Default PMD settings. All probes are trained on a subset of \num{160} YouTube-VOS~\citep{youtubevos} videos. For Depth Anything 3 (DA3), we use the DA3-GIANT-1.1 model.}
\label{tab:pmd_default}
\begin{tabularx}{\textwidth}{@{}l >{\raggedright\arraybackslash}X c c@{}}
\toprule
Backbone & Selected Blocks & Probe & Rank \(k\) \\
\midrule
$\pi^3$/$\pi^3$X~\citep{pi3} & \num{18}\,\texttimes\, inter-frame blocks & low-rank quad. & 8 \\
DA3~\citep{da3} & \num{14}\,\texttimes\, inter-frame blocks & low-rank quad. & 8 \\
DUSt3R~\citep{dust3r} & \num{12}\,\texttimes\, decoder-view-1 + \num{4}\,\texttimes\, head-in-view-1 blocks & low-rank quad. & 8 \\
\bottomrule
\end{tabularx}
\end{table}

\subsection{Zero-shot Cross-Benchmark Motion Segmentation}
\label{sec:generalization}

\begin{table}[H]
\centering
\definecolor{rankgreen}{HTML}{287A46}
\providecommand{\rankfirst}[1]{\begingroup\setlength{\fboxsep}{1.2pt}\colorbox{rankgreen!38}{\textbf{#1}}\endgroup}
\providecommand{\ranksecond}[1]{\begingroup\setlength{\fboxsep}{1.2pt}\colorbox{rankgreen!23}{#1}\endgroup}
\providecommand{\rankthird}[1]{\begingroup\setlength{\fboxsep}{1.2pt}\colorbox{rankgreen!11}{#1}\endgroup}
\caption{Motion segmentation benchmarking. Cells report J-M/J-R ($\uparrow$). \rankfirst{First}, \ranksecond{second}, and \rankthird{third} mark the top three distinct reported scores for each dataset and metric across all rows; ties share a color. Non-PMD scores are quoted from the cited papers; PMD scores are evaluated by us. Gray text marks target-benchmark fine-tuning; ``--'' denotes an unreported score.}
\label{tab:sota_full}
\small
\setlength{\tabcolsep}{3pt}
\renewcommand{\arraystretch}{1.08}
\begin{tabularx}{\textwidth}{@{}l >{\raggedright\arraybackslash}X c c c@{}}
\toprule
Method & Motion-mask supervision; inference components & \makecell{DAVIS17\\J-M / J-R} & \makecell{SegTrackv2\\J-M / J-R} & \makecell{FBMS-59\\J-M / J-R} \\
\midrule
\rowcolor{gray!10}\multicolumn{5}{l}{\textit{Baselines without motion-mask supervision}} \\
VGGT4D~\citep{vggt4d} & None; VGGT geometry & 56.5/65.6 & -- & -- \\
Easi3R\textsubscript{dust3r}~\citep{easi3r} & None; DUSt3R geometry & 46.9/50.5 & -- & -- \\
Easi3R\textsubscript{dust3r}+SAM2 & None; DUSt3R geometry, SAM2 & 60.1/65.3 & -- & -- \\
\midrule
\rowcolor{gray!10}\multicolumn{5}{l}{\textit{Baselines with motion-mask supervision}} \\
GeoMotion~\citep{geomotion} & Five datasets; $\pi^3$, RAFT, DINOv2 & \rankthird{75.4}/87.2 & -- & -- \\
GeoMotion+SAM2 & Five datasets; $\pi^3$, RAFT, DINOv2, SAM2 & \ranksecond{81.1}/\rankfirst{92.6} & \rankthird{77.3}/-- & 72.5/-- \\
Flow-SAM~\citep{flow_sam} & Synthetic masks, DAVIS fine-tuning; RAFT, SAM & \textcolor{gray}{71.6}/-- & \ranksecond{84.2}/-- & \rankfirst{88.7}/-- \\
SegAnyMo~\citep{seg_any_mo} & Three datasets; tracks, depth, DINOv2, SAM2 & -- & 76.3/-- & \rankthird{78.3}/-- \\
\midrule
\rowcolor{gray!10}\multicolumn{5}{l}{\textbf{\textit{Ours: low-rank supervised probe}}} \\
 DUSt3R-PMD & 160 YouTube-VOS videos; DUSt3R geometry & 62.2/70.0 & 58.6/77.3 & 64.0/74.5 \\
DA3-PMD & 160 YouTube-VOS videos; DA3 geometry & 69.6/85.8 & 66.3/80.3 & 69.8/81.4 \\
$\pi^3$-PMD & 160 YouTube-VOS videos; $\pi^3$ geometry & 72.7/89.9 & 72.9/\rankthird{92.9} & 72.6/\rankthird{83.4} \\
$\pi^3$X-PMD & 160 YouTube-VOS videos; $\pi^3$X geometry & 73.7/\rankthird{91.0} & 72.9/\ranksecond{94.0} & 73.5/\ranksecond{88.8} \\
\cmidrule(lr){1-5}
$\pi^3$X-PMD+SAM2 & 160 YouTube-VOS videos; $\pi^3$X geometry, SAM2 & \rankfirst{81.9}/\ranksecond{92.3} & \rankfirst{86.4}/\rankfirst{99.3} & \ranksecond{83.2}/\rankfirst{89.3} \\
\bottomrule
\end{tabularx}
\end{table}

As shown in Table~\ref{tab:sota_full}, PMD generalizes zero-shot from YouTube-VOS to all three test benchmarks without target-domain fine-tuning. Its multi-view pretrained variants achieve stronger performance than DUSt3R-PMD. In particular, $\pi^3$- and $\pi^3$X-PMD surpass Easi3R~\citep{easi3r} and VGGT4D~\citep{vggt4d} by a clear margin, while remaining competitive with supervised methods~\citep{geomotion,flow_sam,seg_any_mo}. The following analyses focus on $\pi^3$X-PMD, which achieves the strongest overall PMD results. Additional benchmark results, including DAVIS16 and DAVIS-all, are provided in Appendix~\ref{app:full_benchmark}.

\paragraph{Coarse Masks and Optional Refinement.}
On DAVIS17, coarse $\pi^3$X-PMD reaches 73.7 J-M and 91.0 J-R, compared with 75.4/87.2 for GeoMotion. SAM2 refinement improves $\pi^3$X-PMD to 81.9/92.3. The J-R gains are 1.2/1.3 points for $\pi^3$/$\pi^3$X-PMD, 5.4 for GeoMotion, and 9.9--14.8 for Easi3R (Table~\ref{tab:sam2_refinement_mini}). High coarse J-R indicates that most frames already exceed 0.5 Jaccard, while the J-M increase reflects improved average mask overlap after refinement.

\begin{table}[!htbp]
\centering
\caption{SAM2~\citep{sam2} refinement on DAVIS2017~\citep{davis17}. $\Delta$J-R is the percentage-point increase in the fraction of frames with Jaccard above 0.5. Additional backbones and methods are reported in Appendices~\ref{app:additional_backbone_benchmarking} and~\ref{app:ff_geometry_comparison}.}
\label{tab:sam2_refinement_mini}
\setlength{\tabcolsep}{5pt}
\begin{tabular}{l c cc cc c}
\toprule
\multirow{2}{*}{Method} 
& \multirow{2}{*}{Flow} 
& \multicolumn{2}{c}{w/o SAM2} 
& \multicolumn{2}{c}{w/ SAM2}
& \multirow{2}{*}{$\Delta$J-R $\downarrow$} \\
\cmidrule(lr){3-4} \cmidrule(lr){5-6}
& 
& J-M $\uparrow$ 
& J-R $\uparrow$ 
& J-M $\uparrow$ 
& J-R $\uparrow$
& \\
\midrule
Easi3R\textsubscript{dust3r}~\citep{easi3r}   
& \xmark 
& 46.9 
& 50.5 
& 60.1 
& 65.3
& 14.8 \\

Easi3R\textsubscript{monst3r}~\citep{easi3r}  
& \xmark 
& 54.8 
& 66.2 
& 67.9 
& 76.1
& 9.9 \\

GeoMotion~\citep{geomotion}                   
& \cmark 
& \textbf{75.4} 
& 87.2 
& 81.1 
& \textbf{92.6}
& 5.4 \\

$\pi^3$-PMD (ours)                                 
& \xmark 
& 72.7 
& 89.9 
& 80.2 
& 91.1
& \textbf{1.2} \\

$\pi^3$X-PMD (ours)                                 
& \xmark 
& 73.7 
& \textbf{91.0} 
& \textbf{81.9} 
& 92.3
& 1.3 \\
\bottomrule
\end{tabular}
\end{table}

\paragraph{Backbone-dependent Recoverability.}
Performance also varies among multi-view backbones: the $\pi^3$X--DA3 gap on DAVIS17 is 4.1 J-M for coarse masks and 4.2 points after SAM2 refinement. Source-validation pairwise decodability is positively associated with mean target coarse J-M across six backbones. Appendix~\ref{app:backbone_disparity} reports the block-level statistics, correlations, and broad design regimes.

\begin{figure}[!htbp]
    \centering
    \includegraphics[width=\linewidth]{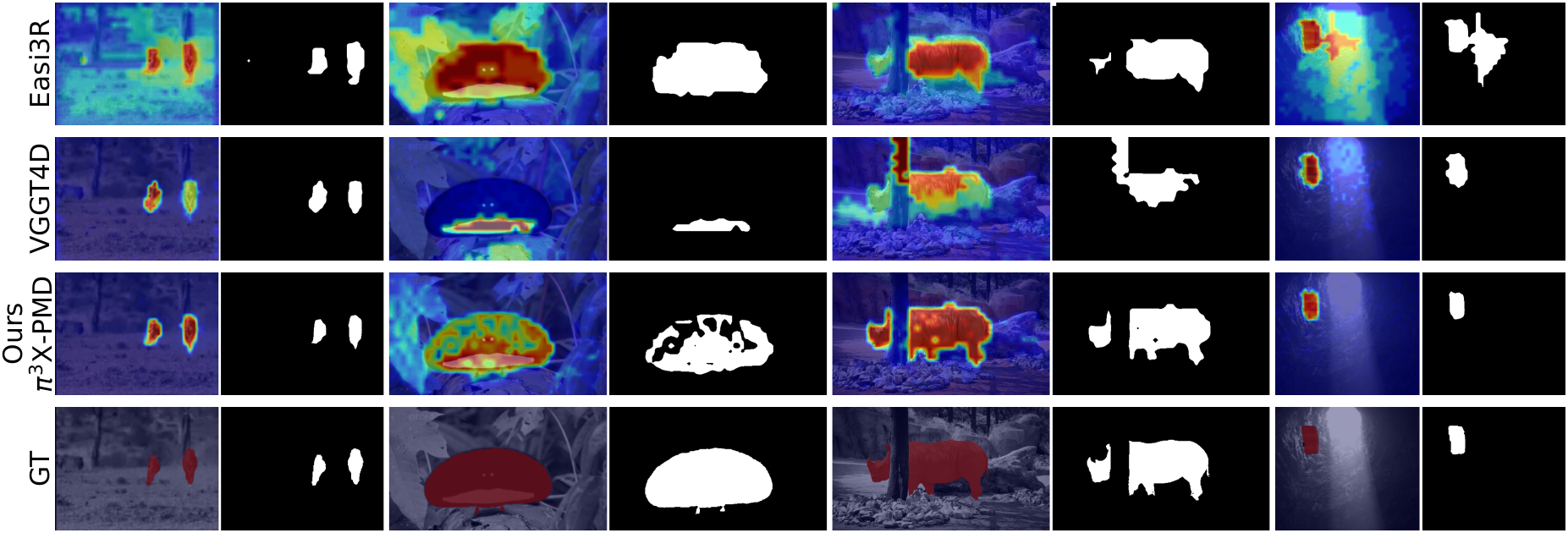}
    \caption{Qualitative comparison on DAVIS~\citep{davis16} and SegTrackv2~\citep{segtrack} against training-free methods based on feed-forward 3D foundation models. The semantic classes shown here are absent from the 160-video training subset of $\pi^3$X-PMD. All binary masks shown here are generated without SAM2~\citep{sam2} refinement. PMD produces cleaner dynamic score maps and more coherent coarse masks.}
    \label{fig:unseen_class_motion_comparison}
\end{figure}

\paragraph{Generalization to Unseen Semantic Categories.}
Figure~\ref{fig:unseen_class_motion_comparison} shows selected examples from DAVIS and SegTrackv2 whose semantic categories are absent from the 160-video YouTube-VOS~\citep{youtubevos} training subset. PMD produces cleaner dynamic score maps and more coherent coarse masks than Easi3R~\citep{easi3r} and VGGT4D~\citep{vggt4d}, illustrating transfer to categories unseen during probe training.

\begin{samepage}
\paragraph{Efficiency.}
Table~\ref{tab:unified_efficiency} compares training data, trainable capacity, and inference overhead. Trained on only 160 videos, $\pi^3$X-PMD uses a 0.295M-parameter low-rank probe and adds 1.62G FLOPs per frame for scoring and aggregation. On a single RTX 5070 Ti, coarse PMD adds 4.2\,ms/frame (0.87\%) to frozen-backbone inference. End-to-end latency and peak VRAM, including the backbone and optional SAM2 refinement, are reported in Appendix~\ref{app:efficiency}.
\par
\end{samepage}

\begin{table*}[!htbp]
\centering
\caption{
Comparison of data efficiency, performance, and inference overhead.
Mask scores use the complete pipelines, including SAM/SAM2 where applicable; PMD and GeoMotion scores include SAM2. Trainable Params count optimized parameters only. Added FLOPs exclude SAM/SAM2 and, for PMD, frozen-backbone inference; end-to-end latency and measurement details are reported in Appendix~\ref{app:efficiency}.
}
\label{tab:unified_efficiency}
\small 
\renewcommand{\arraystretch}{1.15}
\setlength{\tabcolsep}{4pt} 
\renewcommand{\tabularxcolumn}[1]{m{#1}}

\begin{tabularx}{\textwidth}{@{} l >{\raggedright\arraybackslash}X c c c c c @{}}
\toprule
\makecell[l]{Method}
& \makecell[l]{Training Data}
& \makecell[c]{DAVIS17 \\ (J-M $\uparrow$)}
& \makecell[c]{SegTrackv2 \\ (J-M $\uparrow$)}
& \makecell[c]{FBMS-59 \\ (J-M $\uparrow$)}
& \makecell[c]{Trainable \\ Params (M) $\downarrow$}
& \makecell[c]{Added \\ FLOPs (G) $\downarrow$} \\
\midrule

\rowcolor{gray!10} \multicolumn{7}{l}{\textit{Large-scale supervised}} \\
SegAnyMo~\citep{seg_any_mo}
& Kubric + \newline Dynamic Replica + \newline HOI4D + etc.
& -- & \num{76.3} & \num{78.3} & \num{3.44} & \textgreater{}\num{2.08} \\
\arrayrulecolor{gray!30}\midrule\arrayrulecolor{black} 

Flow-SAM~\citep{flow_sam} 
& Synthetic pretraining + \newline DAVIS fine tuning + etc.
& \num{71.6} & \num{84.2} & \textbf{\num{88.7}} & \num{9.83} 
& \textgreater{}\num{3.63e4} \\
\arrayrulecolor{gray!30}\midrule\arrayrulecolor{black} 

GeoMotion~\citep{geomotion}
& GOT-Motion + \newline HOI4D + Dynamic Replica + etc.
& \num{81.1} & \num{77.3} & \num{72.5} & \num{80.3} 
& \textgreater{}\num{3.09e2} \\

\midrule
\rowcolor{gray!10} \multicolumn{7}{l}{\textbf{\textit{Ours (data-efficient probing)}}} \\
$\pi^3$-PMD
& \num{160} YouTube-VOS 2018 videos
& \num{80.2} & \num{84.5} & \num{81.1} & \textbf{\num{0.295}} & \textbf{\num{1.62}} \\
\arrayrulecolor{gray!30}\midrule\arrayrulecolor{black}

$\pi^3$X-PMD
& \num{160} YouTube-VOS 2018 videos
& \textbf{\num{81.9}} & \textbf{\num{86.4}} & \num{83.2} & \textbf{\num{0.295}} & \textbf{\num{1.62}} \\
\bottomrule
\end{tabularx}
\end{table*}

\subsection{Compact Static--Dynamic Discriminative Subspace and Parsimonious Decoding}
\label{sec:mechanism}

The strong zero-shot performance of $\pi^3$X-PMD, despite its small probe capacity and limited training data, suggests that motion cues in its frozen token representations are structured enough to be projected into a compact static--dynamic discriminative subspace.

\begin{figure}[!htbp]
    \centering
    \begin{subfigure}{0.32\linewidth}
        \centering
        \includegraphics[width=\linewidth]{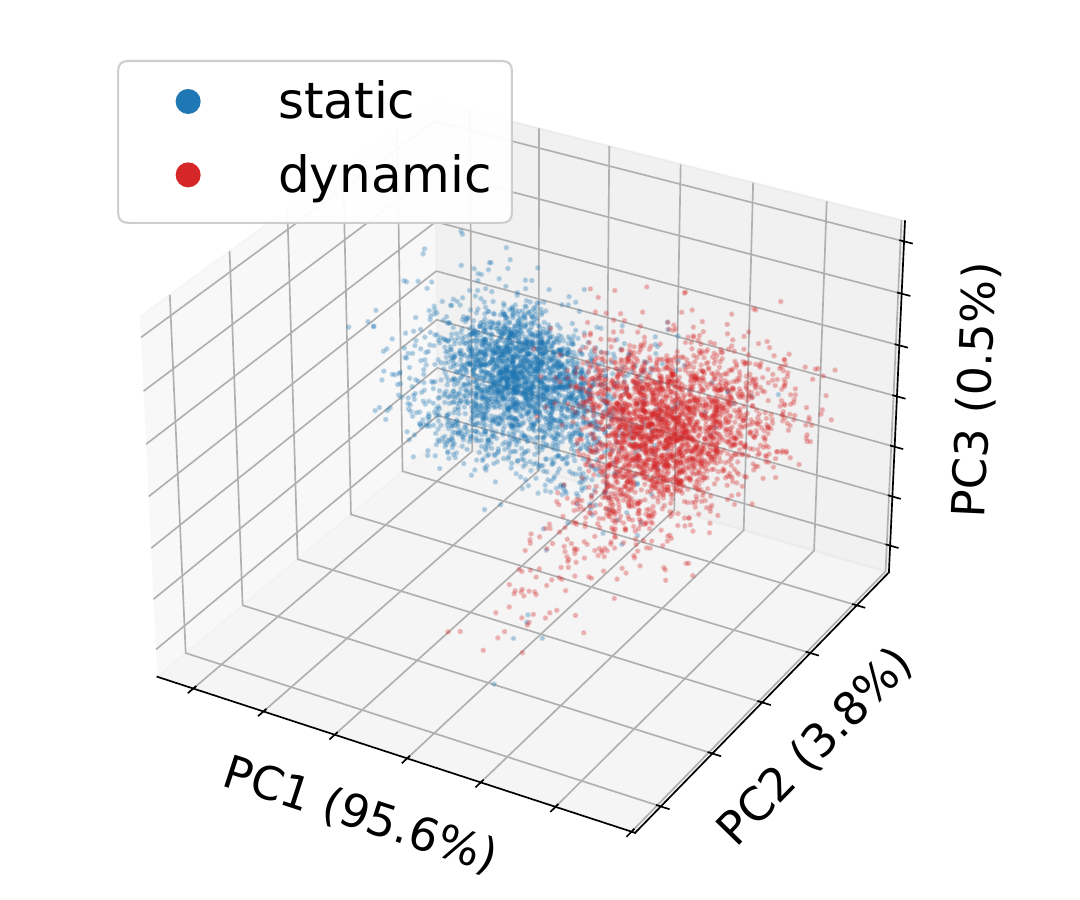}
        \caption{}
        \label{fig:pca_analysis}
    \end{subfigure}
    \hfill
    \begin{subfigure}{0.32\linewidth}
        \centering
        \includegraphics[width=\linewidth]{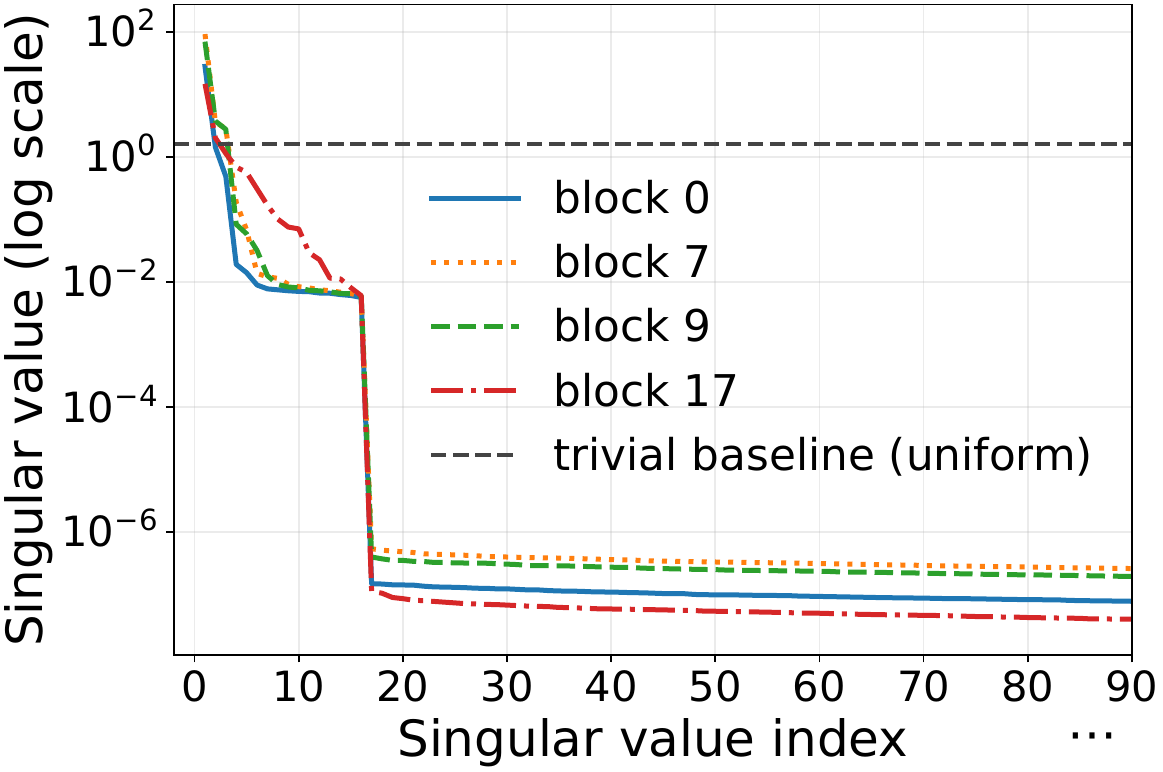}
        \caption{}
        \label{fig:singular_spectrum}
    \end{subfigure}
    \hfill
    \begin{subfigure}{0.32\linewidth}
        \centering
        \includegraphics[width=\linewidth]{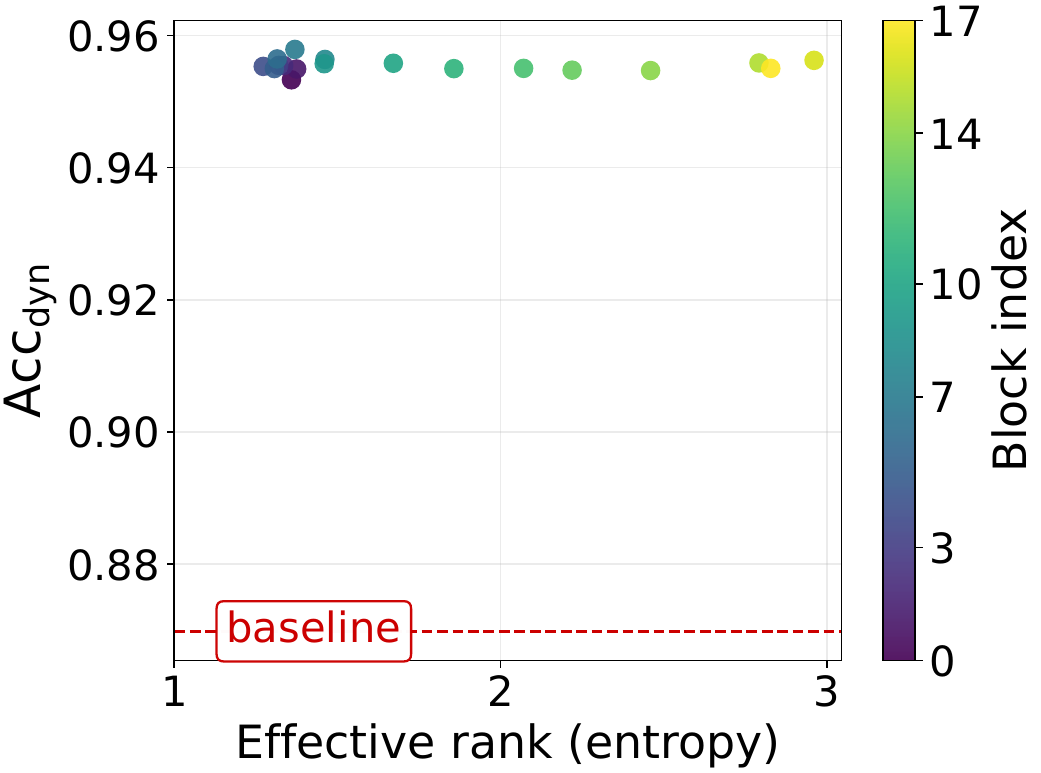}
        \caption{}
        \label{fig:metric_rank}
    \end{subfigure}
    \caption{Low-rank structure of the probe-induced static--dynamic discriminative representation on $\pi^3$X-PMD, illustrated for DAVIS-all~\citep{davis17}. (a) Principal Component Analysis (PCA) illustrates clear separation between static and dynamic tokens. (b) The singular spectrum decays steeply in the leading components, especially in earlier layers, revealing strong rank concentration. (c) Performance saturates at a very low effective rank, indicating that the probe-induced motion cues are compact.}
    \label{fig:lowrank_structure}
\end{figure}

Figure~\ref{fig:lowrank_structure} provides a qualitative and quantitative analysis of this probe-induced subspace. In Figure~\ref{fig:pca_analysis}, a 3D PCA visualization of the low-rank quadratic probe representation (\(k=8\)) exhibits a clear separation between static and dynamic tokens. Figure~\ref{fig:singular_spectrum} further shows rapid spectral decay, indicating that the probe-induced motion cues are concentrated in a small number of dominant directions. This spectral concentration is also reflected in Figure~\ref{fig:metric_rank}, where strong performance is retained even when the probe-induced motion cues have low effective rank, particularly in shallow layers.

\begin{figure}[!htbp]
    \centering
    \begin{subfigure}{\linewidth}
        \centering
        \includegraphics[width=\linewidth]{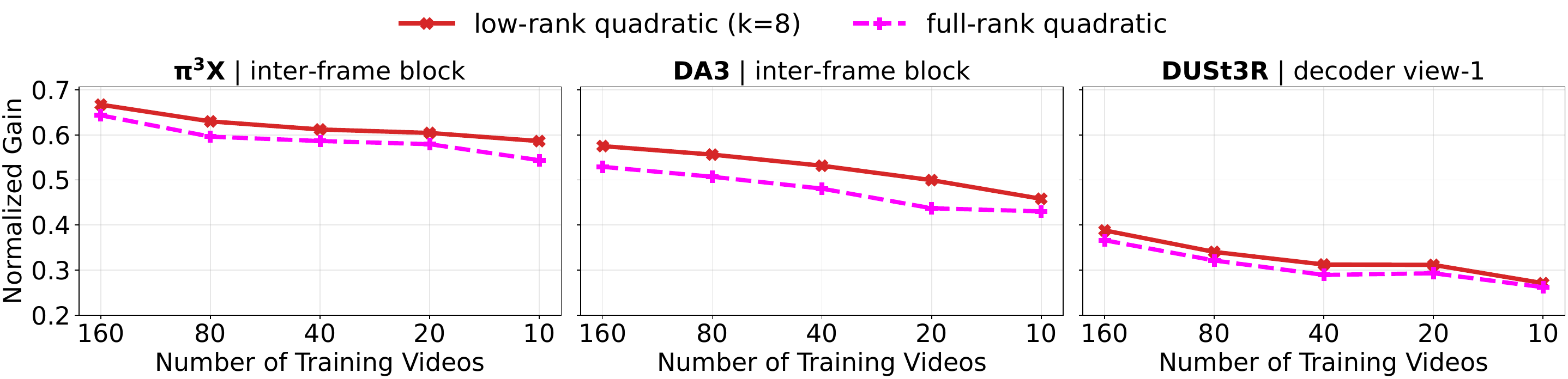}
        \caption{Token-level probing (Normalized Gain) on DAVIS-all~\citep{davis17}.}
        \label{fig:data_starvation_ng}
    \end{subfigure}
    
    \vspace{2pt}
    
    \begin{subfigure}{\linewidth}
        \centering
        \includegraphics[width=\linewidth]{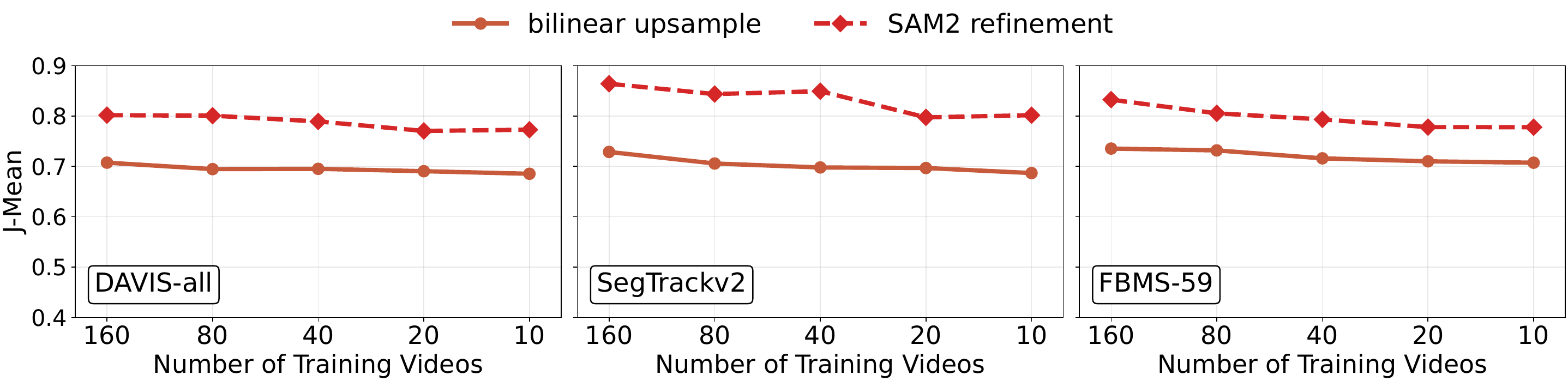}
        \caption{Downstream mask prediction performance (J-M) before and after SAM2~\citep{sam2} refinement.}
        \label{fig:data_starvation_j_mean}
    \end{subfigure}
    
    \caption{
    Scene-diversity starvation analysis. 
    (a) Low-rank probing ($k=8$) matches or exceeds full-rank probing under reduced training data, indicating a compact static--dynamic discriminative structure. 
    (b) Downstream mask quality degrades gradually rather than collapsing.
    }
    \label{fig:data_starvation}
\end{figure}

We next test whether the compact motion cues remain stable under reduced scene diversity. Keeping the number of sampled token pairs fixed, we progressively reduce the number of training videos. As shown in Figure~\ref{fig:data_starvation_ng}, normalized gain degrades gradually, while the low-rank probe (\(k=8\)) consistently matches or exceeds the full-rank probe. This supports low rank as an effective inductive bias for the supervised readout. Figure~\ref{fig:data_starvation_j_mean} shows a similar trend at the mask level: downstream performance decreases gradually, both before and after SAM2~\citep{sam2} refinement.

\paragraph{Probe Selectivity.}
To test whether the probe relies on token--motion-state correspondence, we shuffle labels within each frame group and retrain the same rank-8 $\pi^3$X probe, preserving frozen features and label counts. On YouTube-VOS validation, 84.47\% of pairs share the same motion state, so always predicting ``same state'' already attains this accuracy. Balanced accuracy averages recall for same-state and different-state pairs equally; this constant predictor scores 50\%. Averaged over 18 inter-frame blocks, shuffling lowers pair accuracy from 95.72\% to 84.48\% and balanced accuracy from 91.67\% to 50.34\% (Table~\ref{tab:probe_selectivity}). The return to baseline indicates that the gains rely on the correct token--label correspondence. Details are in Appendix~\ref{app:selectivity}.

\begin{table}[!htbp]
\centering
\caption{Selectivity of low-rank quadratic probes on frozen $\pi^3$X tokens. Gain is the percentage-point improvement in pair accuracy over the always-same-state baseline.}
\label{tab:probe_selectivity}
\small
\setlength{\tabcolsep}{5pt}
\begin{tabular}{lcccc}
\toprule
Probe / labels & Params & \makecell{Pair accuracy\\(\%)} & \makecell{Gain\\(pp)} & \makecell{Balanced accuracy\\(\%)} \\
\midrule
$k=4$, correct & 0.147M & 95.70 & +11.23 & 91.60 \\
$k=8$, correct & 0.295M & 95.72 & +11.25 & 91.67 \\
$k=8$, shuffled & 0.295M & 84.48 & +0.01 & 50.34 \\
Always-same baseline & 0 & 84.47 & 0.00 & 50.00 \\
\bottomrule
\end{tabular}
\end{table}

\section{Limitations}
\label{sec:limitations}

\begin{figure}[!htbp]
    \centering
    \begin{subfigure}{0.49\linewidth}
        \centering
        \includegraphics[width=\linewidth]{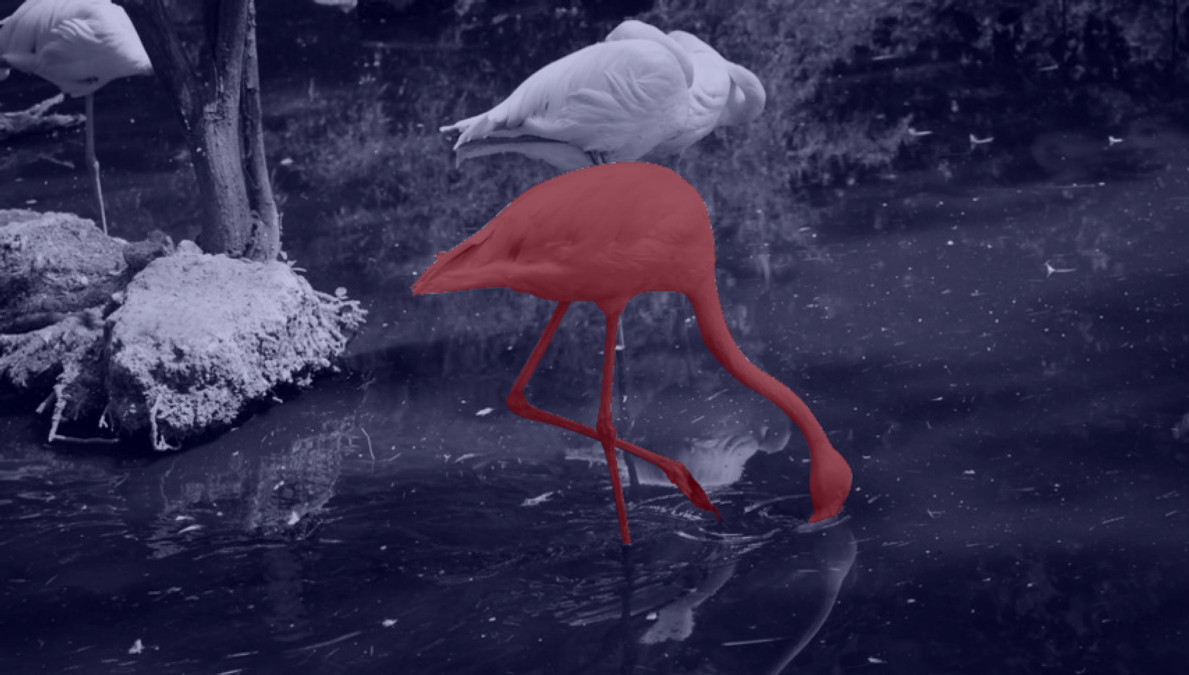}
        \caption{Ground truth}
    \end{subfigure}
    \hfill
    \begin{subfigure}{0.49\linewidth}
        \centering
        \includegraphics[width=\linewidth]{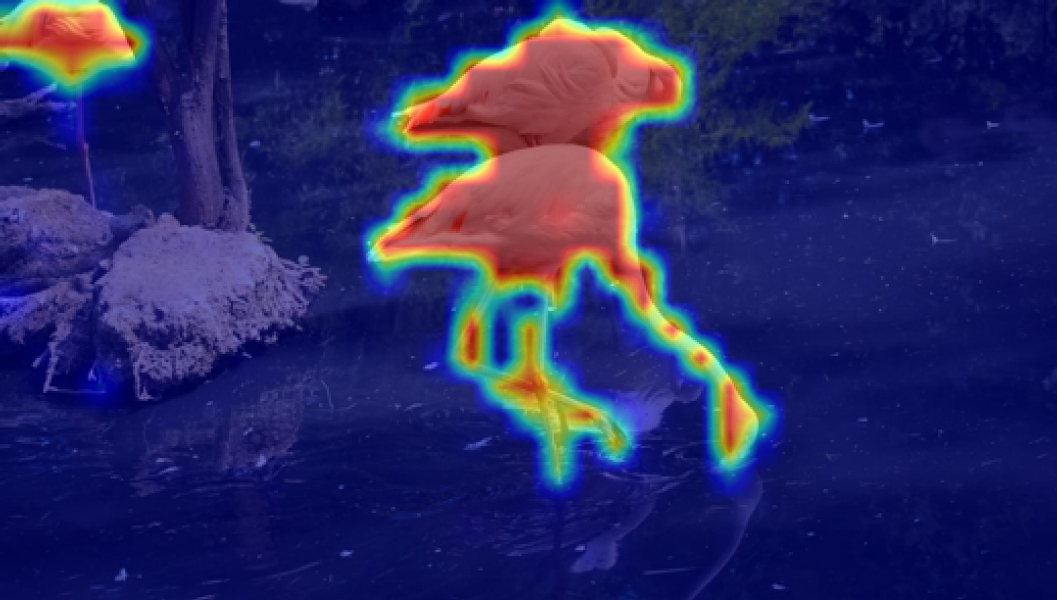}
        \caption{PMD dynamic score map}
    \end{subfigure}
    \caption{Semantic leakage in $\pi^3$X-PMD. The two static background flamingos receive elevated dynamic scores alongside the moving foreground instance, illustrating imperfect separation of motion and semantic cues.}
    \label{fig:app_limitation_leakage}
\end{figure}

While PMD shows strong cross-benchmark generalization and data efficiency, its spatial precision is limited by ViT token granularity. PMD makes binary static--dynamic decisions and does not distinguish fine-grained motion types such as in-place articulation, local deformation, or object displacement. In some cases, PMD also exhibits semantic leakage (Figure~\ref{fig:app_limitation_leakage}): static instances may receive elevated dynamic scores when another instance of the same category is moving, even for categories absent from the 160-video training subset. Object-centric YouTube-VOS supervision and frozen semantic features may contribute to this behavior. Examples are provided in Appendix~\ref{app:limitations}. Future work could probe additional physical quantities, such as displacement and velocity, from frozen 3D foundation models to distinguish different motion modes.

\section{Conclusion}
In this work, we show that frozen feed-forward 3D foundation models already contain motion cues that can be decoded without high-capacity motion decoders. Using $\pi^3$X~\citep{pi3}, DA3~\citep{da3}, and DUSt3R~\citep{dust3r} as representative backbones, we find that lightweight low-rank quadratic probes can expose static--dynamic discriminative cues from frozen token representations. For $\pi^3$X~\citep{pi3}, our rank-restricted and spectral analyses further show that these cues are concentrated in a compact low-dimensional discriminative subspace. When trained on only 160 YouTube-VOS~\citep{youtubevos} videos, our Probing-based Motion Distillation (PMD) transfers zero-shot to DAVIS~\citep{davis16,davis17}, SegTrackv2~\citep{segtrack}, and FBMS-59~\citep{fbms} while remaining competitive with substantially heavier methods.

These results suggest that motion understanding in multi-view pretrained 3D foundation models can be treated as a lightweight decoding problem of frozen tokens rather than re-learning motion-specific features through large-scale training. This perspective motivates probing emerging capabilities of 3D foundation models before resorting to large decoders, extensive supervision, or task-specific adaptation.

\begin{ack}
We thank the anonymous reviewers and the area chair for their
constructive feedback.

\textbf{Funding statement.}
This research received no external funding.

\textbf{Competing interests.}
The authors declare no competing interests.
\end{ack}

\bibliographystyle{plainnat}
\bibliography{references}

\clearpage
\appendix

\section{Implementation Details}
\label{app:implementation}

\paragraph{Hardware.}
All experiments were run on a single workstation with an NVIDIA GeForce RTX 5070 Ti GPU (16\,GB memory), an AMD Ryzen 7 7700X CPU, and 64\,GB of system memory. Probe training, motion-mask extraction, and SAM2~\citep{sam2} refinement were performed on one GPU at a time; no distributed training was used.

\paragraph{Training and Validation Data.}
Unless otherwise specified, probes are trained on 160 videos sampled from the official YouTube-VOS 2018~\citep{youtubevos} training split and validated on 80 videos sampled from its validation split. No videos or annotations from DAVIS~\citep{davis16, davis17}, SegTrackv2~\citep{segtrack}, or FBMS-59~\citep{fbms} are used to optimize probe training parameters.

\paragraph{Training Hyperparameters.}
We optimize the probe weights using AdamW with a constant learning rate of \(1\times 10^{-3}\). We do not use weight decay or learning-rate scheduling. All probes are trained for 16 epochs. For each epoch, we sample 200K token pairs from within the same exported token group, i.e., from tokens jointly sampled from the same target frame.

\paragraph{Inference Context.}
At inference time, PMD is applied to target-frame tokens extracted from a multi-frame input group. The surrounding frames provide temporal or multi-view context to the frozen backbone, while probing and mask decoding are performed on the target frame. In all the motion-mask extraction experiments, we use a 3-frame context with stride 2.

\paragraph{Mask Extraction.}
After obtaining token-wise dynamic score maps, we bilinearly upsample them to image resolution. Coarse PMD uses Otsu thresholding with an offset of 0.2. PMD+SAM2 constructs a prompt mask with the same threshold, erodes it for two iterations only during prompt construction, and filters connected components by their mean dynamic score using Otsu with an offset of 0.3 before SAM2~\citep{sam2} refinement. These extra operations belong to the optional refinement pipeline.

\paragraph{Threshold Robustness.}
Table~\ref{tab:threshold_ablation_coarse_pmd} evaluates nine binarization rules without SAM2. Each rule is applied unchanged across all sequences and benchmarks, without sequence- or dataset-specific selection. For fixed thresholds 0.40--0.60 and Otsu offsets 0.0--0.2, the across-rule standard deviation of mean J-M is 0.69 and 0.97, respectively. Otsu+0.1 attains a higher mean than the Otsu+0.2 default (73.0 versus 72.4); thus, Otsu+0.2 is not uniquely privileged. We retain it as the common default. Extreme thresholds reduce performance, and benchmark averages do not separately establish robustness to illumination or clutter.

\begin{table}[!htbp]
\centering
\caption{Threshold robustness of coarse $\pi^3$X-PMD without SAM2. Mean and Std. summarize J-M across the three benchmarks. The last two rows instead report across-rule standard deviations for each benchmark and for the three-benchmark mean.}
\label{tab:threshold_ablation_coarse_pmd}
\small
\setlength{\tabcolsep}{4pt}
\begin{tabular}{lccccc}
\toprule
Rule & DAVIS-all & SegTrackv2 & FBMS-59 & Mean & Std. \\
\midrule
Fixed 0.30 & 64.5 & 69.4 & 66.6 & 66.8 & 1.98 \\
Fixed 0.40 & 68.1 & 73.3 & 71.1 & 70.8 & 2.15 \\
Fixed 0.50 & 69.9 & 73.3 & 73.9 & 72.4 & 1.75 \\
Fixed 0.60 & 69.7 & 69.4 & \textbf{74.2} & 71.1 & 2.20 \\
Fixed 0.70 & 66.5 & 61.6 & 71.2 & 66.4 & 3.93 \\
Otsu & 69.1 & 72.2 & 70.8 & 70.7 & 1.28 \\
Otsu + 0.1 & \textbf{71.2} & \textbf{74.5} & 73.2 & \textbf{73.0} & 1.34 \\
Otsu + 0.2 (default) & 70.7 & 72.9 & 73.5 & 72.4 & \textbf{1.17} \\
Otsu + 0.3 & 67.5 & 66.2 & 70.8 & 68.2 & 1.93 \\
\midrule
\makecell[l]{Across-rule std.,\\Fixed 0.40--0.60} & 0.81 & 1.84 & 1.40 & 0.69 & -- \\
\makecell[l]{Across-rule std.,\\Otsu + 0.0--0.2} & 0.90 & 0.96 & 1.21 & 0.97 & -- \\
\bottomrule
\end{tabular}
\end{table}

\section{PMD Inference Algorithm}
\begin{algorithm}[H]
\caption{PMD Inference}
\label{alg:pmd_inference}
\begin{algorithmic}[1]
\Require Frozen backbone \(F\), input frame group \(I\), selected blocks \(\mathcal{L}\), trained pairwise probe \(f(\cdot,\cdot)\), static anchor \(a=0\)
\Ensure Dynamic score map \(\hat{\mathbf{P}}^{\mathrm{dyn}}\), coarse mask \(\hat{\mathbf{M}}_{\mathrm{coarse}}\), optional refined mask \(\hat{\mathbf{M}}_{\mathrm{SAM2}}\)

\State Extract target-frame token features \(\mathbf{H}^{(\ell)}=\{\mathbf{h}_0^{(\ell)},\ldots,\mathbf{h}_N^{(\ell)}\}\) for all \(\ell\in\mathcal{L}\).
\For{each selected block \(\ell\in\mathcal{L}\) \textbf{in parallel}}
    \State Compute \(\hat p_{i,\ell}^{\mathrm{stat}}=f(\mathbf{h}_a^{(\ell)},\mathbf{h}_i^{(\ell)})\) for all target-frame tokens \(i\) in parallel.
    \State Convert static probability to dynamic probability via \(\hat p_{i,\ell}^{\mathrm{dyn}}=1-\hat p_{i,\ell}^{\mathrm{stat}}\).
\EndFor
\State Average dynamic scores across blocks.
\State Reshape the averaged token scores to the token grid and bilinearly upsample them to obtain \(\hat{\mathbf{P}}^{\mathrm{dyn}}\).
\State Obtain \(\hat{\mathbf{M}}_{\mathrm{coarse}}\) using Otsu thresholding with offset \(0.2\).

\If{SAM2 refinement is enabled}
    \State Construct a coarse prompt mask using Otsu thresholding with offset \(0.2\).
    \State Erode the prompt mask for two iterations.
    \State Keep connected components whose mean dynamic score passes Otsu thresholding with offset \(0.3\).
    \State Use the surviving components as prompts for SAM2 to obtain \(\hat{\mathbf{M}}_{\mathrm{SAM2}}\).
\EndIf

\State \Return \(\hat{\mathbf{P}}^{\mathrm{dyn}}, \hat{\mathbf{M}}_{\mathrm{coarse}}\), and optionally \(\hat{\mathbf{M}}_{\mathrm{SAM2}}\)
\end{algorithmic}
\end{algorithm}

\section{Dataset Statistics and Fixed-Reference Diagnostics}
\label{app:dataset_statistics}

\subsection{Fixed Reference-Token Convention}
\label{app:convention}
PMD uses the first token \(a=0\) as a fixed reference token throughout all experiments. This choice is fixed a priori and is never selected or adapted per image, video, or benchmark. For padded backbones, including the strongest \(\pi^3\)/\(\pi^3\)X variants in our experiments, this token lies in the padded region and is therefore static by construction. For non-padded backbones such as DA3~\citep{da3} and DUSt3R~\citep{dust3r}, we keep the same fixed-index convention. Table~\ref{tab:fixed_anchor_staticness} shows that this fixed upper-left reference location is static in the vast majority of frames on the training source. These statistics are used only as a post-hoc sanity check and are not used for anchor selection, threshold tuning, model selection, or benchmark-specific adaptation.

Figure~\ref{fig:fixed_anchor_stress} further stress-tests this fixed-reference convention qualitatively on the two non-padded backbones used in our main analysis. We compare a standard case where the fixed anchor patch is static with a rare violation where the same patch is fully dynamic under the ground-truth mask. In the violation case, the PMD score maps show a moderately elevated background response, but the moving foreground remains separable and is recovered by both the coarse and refined masks. This suggests that, for the non-padded backbones evaluated here, PMD is not determined solely by the instantaneous ground-truth state of the reference patch; the fixed spatial index may still provide a useful static-reference prior, likely supported by positional encoding and the highly static distribution of the upper-left patch. Since these two non-padded backbones are included in the main experiments and the stress-test violation does not qualitatively overturn their predictions, we use the same fixed \(a=0\) protocol for all reported results without benchmark-specific anchor correction.

\paragraph{Camera Motion and Uniform Motion States.}
In a fully static scene under camera motion, a uniform near-zero dynamic map is the intended output: image motion alone does not imply independently moving foreground. If all scene tokens share the same motion state, pairwise same-state labels do not distinguish all-static content from uniformly moving content without an independent reference. Under this formulation both admit a no-independent-foreground interpretation; the output is not an estimate of absolute scene motion. No dedicated evaluation on all-static camera-motion datasets is reported here.

Our implementation also provides an optional safeguard that pins \(a=0\) to the static label and explicitly includes anchor--token pairs involving \(a=0\) in each training batch. This safeguard is disabled for all results reported in this paper and is provided only for robustness checks or for applying PMD to datasets where the fixed-anchor convention may not hold.

\begin{table}[!htbp]
\centering
\setlength{\tabcolsep}{4pt}
\renewcommand{\arraystretch}{1.14}
\caption{Fixed-reference staticness diagnostic on YouTube-VOS 2018~\citep{youtubevos}. The last column reports the percentage of frames in which the upper-left \(14\times14\) patch corresponding to the fixed reference token is static under the projected-mask labeling rule.}
\label{tab:fixed_anchor_staticness}
\small
\begin{tabularx}{\textwidth}{@{} l *{4}{>{\centering\arraybackslash}X} @{}}
\toprule
Split
& \makecell{\# Seq.}
& \makecell{\# GT Frames}
& \makecell{Static Pixels (\%)}
& \makecell{Static Upper-left\\\(14{\times}14\) Patch\\(\%)} \\
\midrule

\rowcolor{gray!10} \multicolumn{5}{l}{\textit{Full official split}} \\
Train & 2{,}985 & 89{,}750 & 82.75 & 97.03 \\
Valid & 492 & 16{,}563 & 83.08 & 96.61 \\

\midrule

\rowcolor{gray!10} \multicolumn{5}{l}{\textit{Our subset}} \\
Train & 160 & 4{,}933 & 87.03 & 97.73 \\
Valid & 80 & 2{,}893 & 84.79 & 96.58 \\



\bottomrule
\end{tabularx}
\end{table}

\begin{figure}[!htbp]
    \centering
    \setlength{\tabcolsep}{2pt}
    \renewcommand{\arraystretch}{1.08}
    \small
    \begin{tabular}{@{}
        >{\centering\arraybackslash}m{0.14\linewidth}
        *{5}{>{\centering\arraybackslash}m{0.16\linewidth}}
        @{}}
        &
        \makecell[c]{input frame}
        & GT mask
        & dyn. score map
        & coarse mask
        & refined mask
        \\[1mm]

        \makecell[c]{DA3-PMD\\\(a=0\) stat.}
        &
        \includegraphics[width=\linewidth]{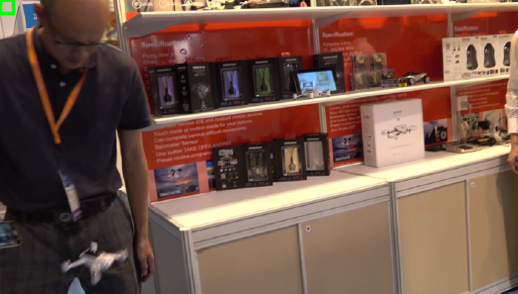}
        &
        \includegraphics[width=\linewidth]{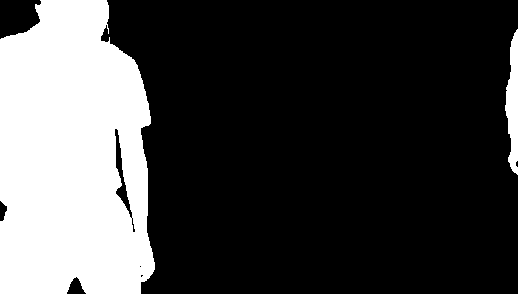}
        &
        \includegraphics[width=\linewidth]{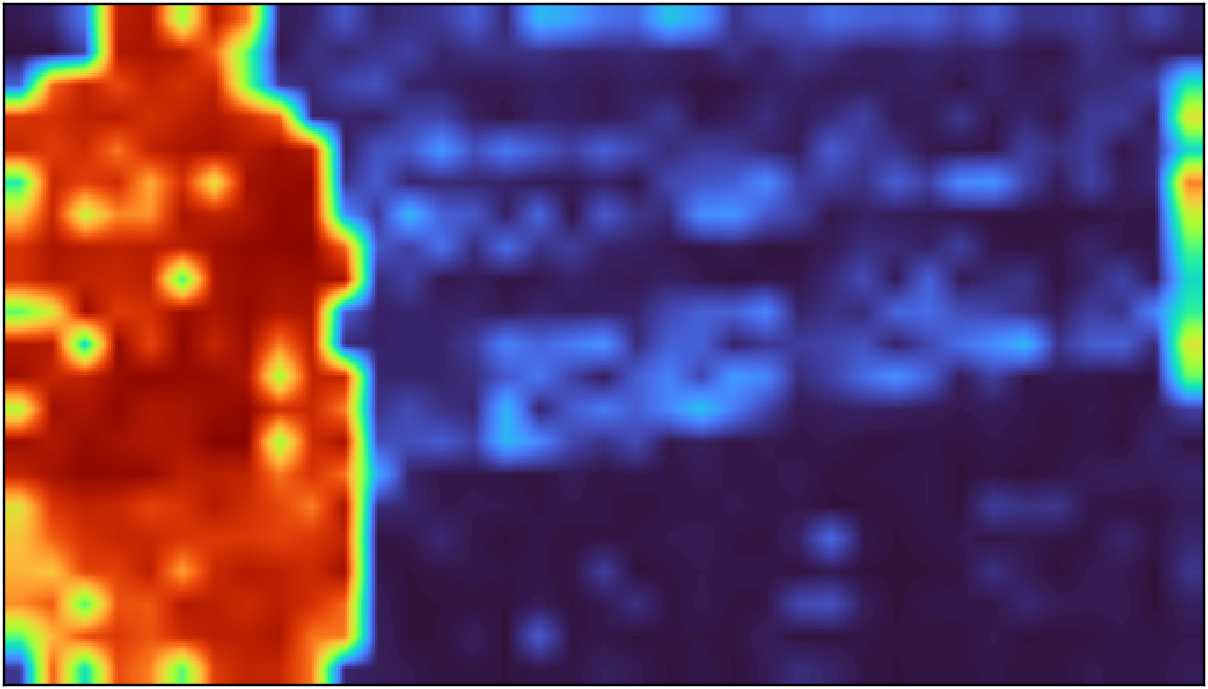}
        &
        \includegraphics[width=\linewidth]{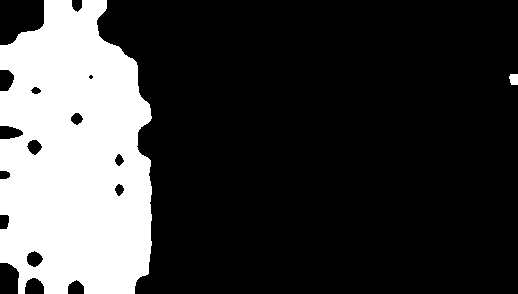}
        &
        \includegraphics[width=\linewidth]{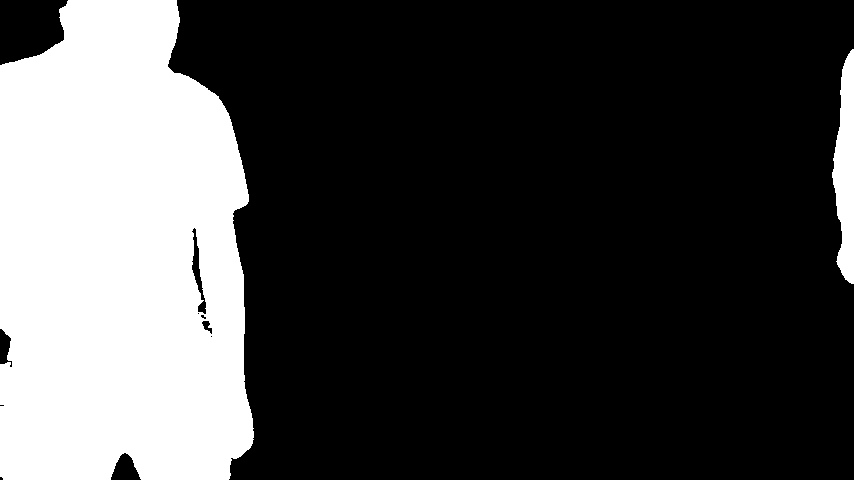}
        \\[1.5mm]

        \makecell[c]{DA3-PMD\\\(a=0\) dyn.}
        &
        \includegraphics[width=\linewidth]{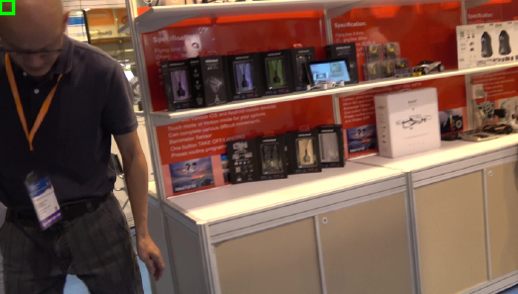}
        &
        \includegraphics[width=\linewidth]{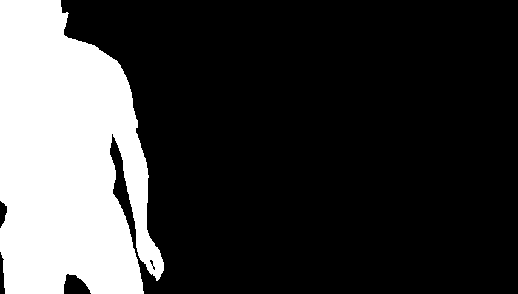}
        &
        \includegraphics[width=\linewidth]{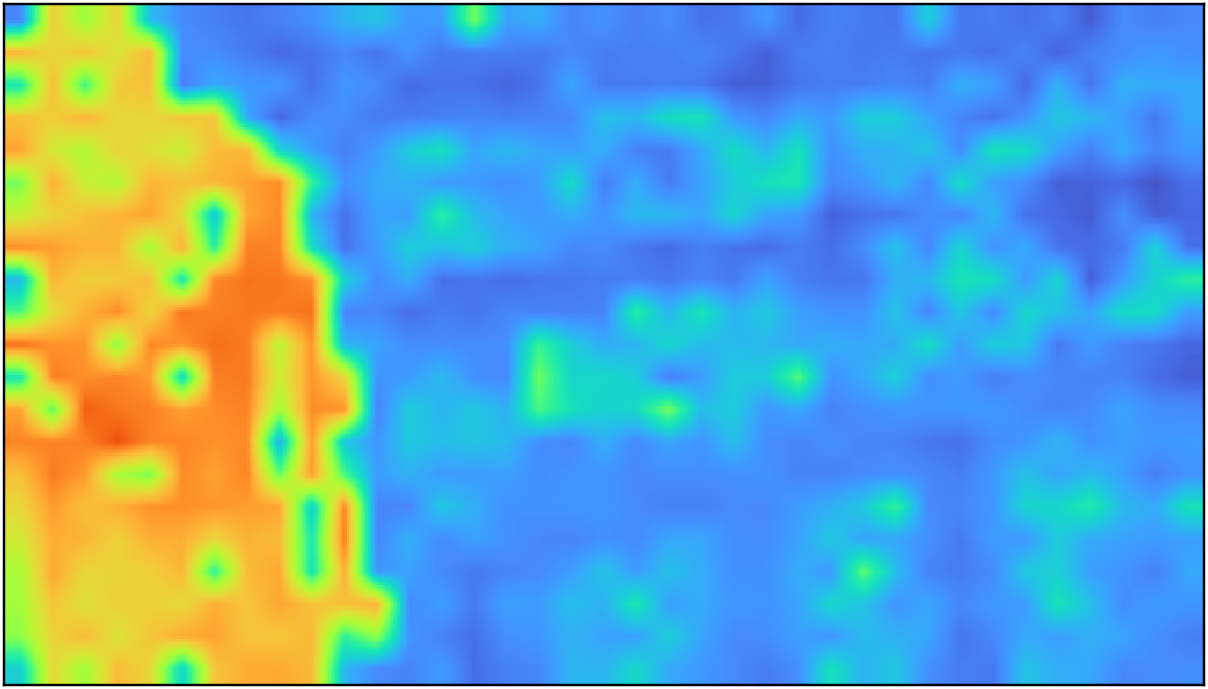}
        &
        \includegraphics[width=\linewidth]{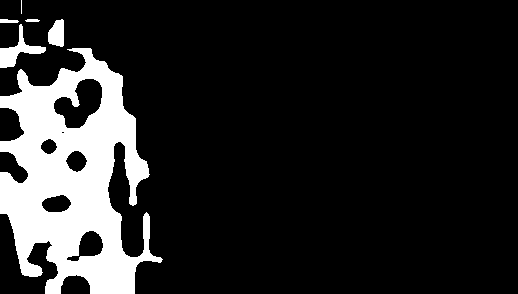}
        &
        \includegraphics[width=\linewidth]{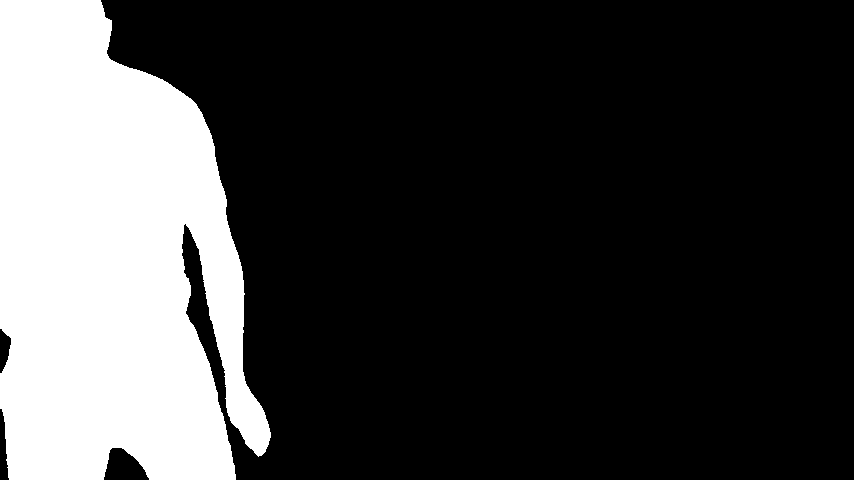}
        \\[2mm]

        \makecell[c]{DUSt3R-PMD\\\(a=0\) stat.}
        &
        \includegraphics[width=\linewidth]{Graphics/DA3/16rgb_patch.png}
        &
        \includegraphics[width=\linewidth]{Graphics/DA3/16_gt_mask.png}
        &
        \includegraphics[width=\linewidth]{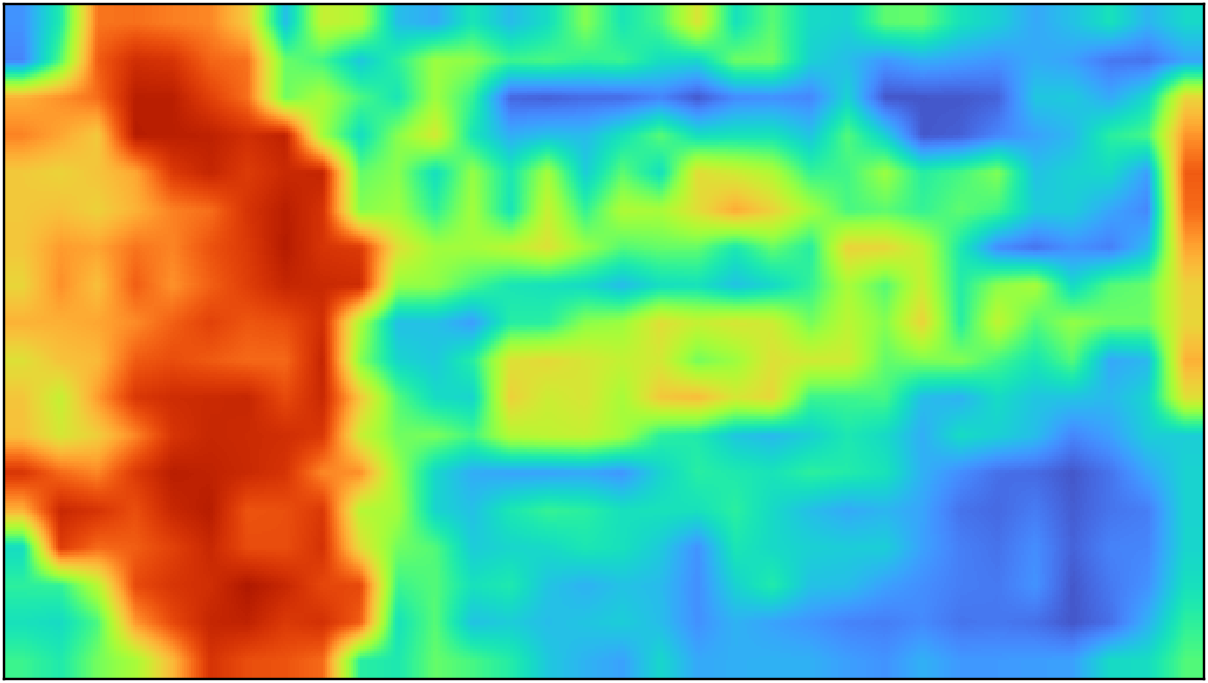}
        &
        \includegraphics[width=\linewidth]{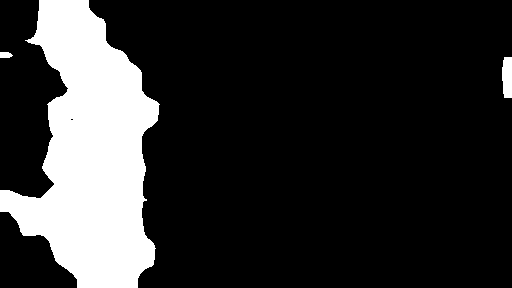}
        &
        \includegraphics[width=\linewidth]{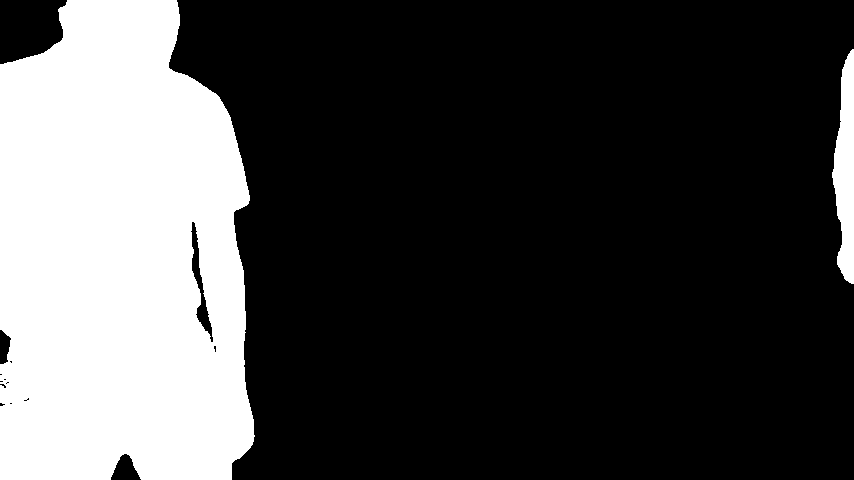}
        \\[1.5mm]

        \makecell[c]{DUSt3R-PMD\\\(a=0\) dyn.}
        &
        \includegraphics[width=\linewidth]{Graphics/DA3/18_rgb_patch.png}
        &
        \includegraphics[width=\linewidth]{Graphics/DA3/18_gt_mask.png}
        &
        \includegraphics[width=\linewidth]{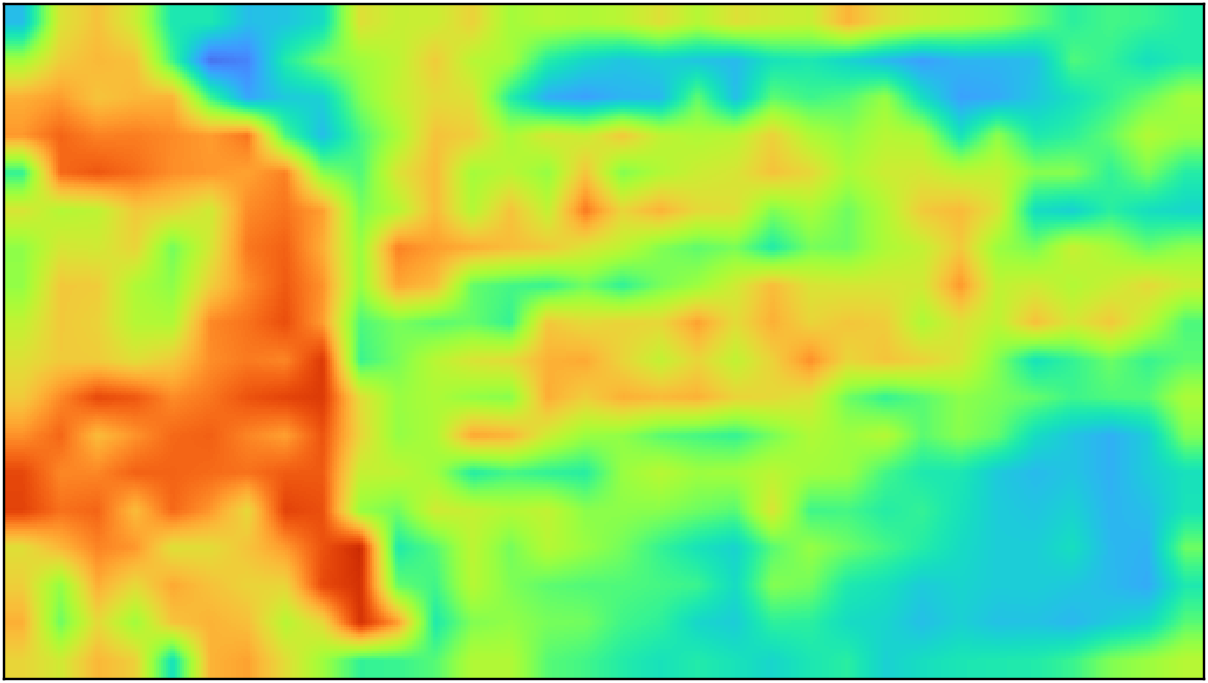}
        &
        \includegraphics[width=\linewidth]{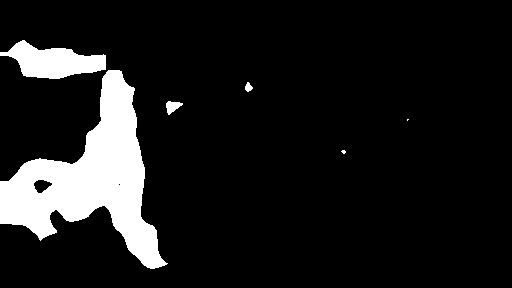}
        &
        \includegraphics[width=\linewidth]{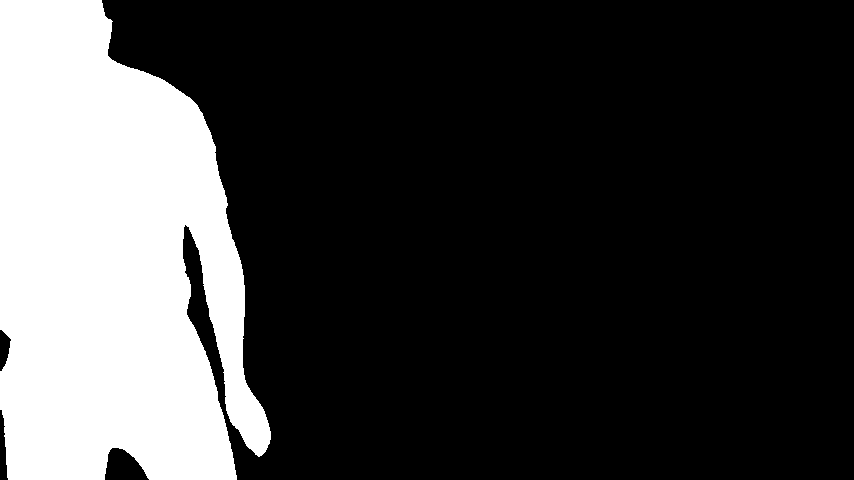}
        \\
    \end{tabular}

    \caption{
    Stress test of the fixed anchor token \(a=0\) convention on a DAVIS-all~\citep{davis17} evaluation example for the two non-padded backbones in the main analysis.
    The upper-left reference patch is marked in green in the input frame.
    For each backbone, we compare a standard case where the fixed \(a=0\) patch is static with a rare violation where the same patch is fully dynamic according to the ground-truth mask.
    For both DA3-PMD and DUSt3R-PMD, the violation moderately increases the background response, but it does not invert the prediction: the moving foreground remains separable and is recovered by both the coarse and refined masks.
    }
    \label{fig:fixed_anchor_stress}
\end{figure}

\subsection{Token-Level Class Imbalance.}
\label{app:imbalance}
In addition to the fixed-anchor diagnostic above, we report token-level static--dynamic imbalance on the evaluation benchmarks in Table~\ref{tab:token_imbalance}. These statistics are used only to contextualize the all-static baseline and the normalized-gain metric in the probing analysis; they are not used for anchor token selection, threshold tuning, or model training.

\begin{table}[!htbp]
\centering
\setlength{\tabcolsep}{4pt}
\caption{Representative token-level class imbalance on the evaluation benchmarks. Static tokens dominate across all test sets. Ratios are computed from the $\pi^3$X~\citep{pi3} tokenization of the probed features; normalized-gain baselines are computed separately for each backbone using its own tokenization.}
\label{tab:token_imbalance}
\begin{tabular}{lccc}
\toprule
Dataset & \# Valid Tokens & Static Tokens (\%) & Dynamic Tokens (\%) \\
\midrule
DAVIS-all~\citep{davis17}   & 392{,}512 & 92.0 & 8.0 \\
SegTrackv2~\citep{segtrack} & 68{,}224  & 95.3 & 4.7 \\
FBMS-59~\citep{fbms}        & 46{,}080  & 88.4 & 11.6 \\
\bottomrule
\end{tabular}
\end{table}

\section{Capacity Analysis of Low-Rank Quadratic Probes}
\label{app:capacity_analysis}

While the main text empirically demonstrates the compactness of the decoded motion representation and PMD's data efficiency, this appendix provides a capacity-based interpretation of why PMD can generalize from a sparse subset of token pairs and only 160 training videos.

\subsection{Pairwise Low-Rank Probe Class}

For a token pair \((i,j)\), PMD predicts the pair label \(z_{ij}\) using the low-rank quadratic probe:
\begin{equation}
\hat z_{ij}^{(k)} = \sigma\!\left(s_{ij}^{(k)}\right),
\end{equation}
with score
\begin{equation}
s_{ij}^{(k)}
=
\frac{1}{2\sqrt d}\,
\mathbf h_i^{(\ell)\top}
\left(
\mathbf U^{(k)\top} \mathbf V^{(k)} + \mathbf V^{(k)\top} \mathbf U^{(k)}
\right)
\mathbf h_j^{(\ell)}
+ b,
\end{equation}
where
\(
\mathbf U^{(k)}, \mathbf V^{(k)} \in \mathbb{R}^{k \times d}
\),
\(b \in \mathbb{R}\), and \(k\) is the rank hyperparameter. Due to the symmetrization, the resulting bilinear matrix has rank at most \(2k\), but its degrees of freedom still scale linearly with \(kd\).

\subsection{Capacity Scaling}

The default $\pi^3$X-PMD configuration averages probe outputs from \(L_{\mathrm{agg}}=18\) selected inter-frame blocks, with one low-rank probe per block. Therefore, the learnable factor matrices contribute
\begin{equation}
P_{\mathrm{PMD}}(k)
=
L_{\mathrm{agg}}(2kd + 1)
=
18(2kd + 1)
\end{equation}
parameters. An unconstrained full-rank quadratic probe over the same aggregated blocks requires
\begin{equation}
P_{\mathrm{full}}
=
L_{\mathrm{agg}}(d^2 + 1)
=
18(d^2 + 1).
\end{equation}
For \(d=1024\), this yields
\begin{equation}
P_{\mathrm{full}}
=
18(1024^2 + 1)
=
18{,}874{,}386.
\end{equation}

\begin{table}[!htbp]
\centering
\caption{Validation-based rank selection for PMD. Parameter counts are reported across candidate rank hyperparameters \(k\), compared with the implemented full-rank quadratic probe at \(d=1024\) and \(L_{\mathrm{agg}}=18\) aggregated blocks. For $\pi^3$X-PMD, we report the best token-level Accuracy and Normalized Gain on the shared YouTube-VOS validation subset (80 sequences) over all inter-frame blocks. This validation subset, not any target evaluation benchmark, is used to select the default rank \(k=8\).}
\label{tab:pmd_capacity_val}
\begin{tabular}{ccccc}
\toprule
Rank \(k\) & \(P_{\mathrm{PMD}}(k)\) & \(P_{\mathrm{PMD}}(k) / P_{\mathrm{full}}\) & Accuracy (\%) $\uparrow$ & NG $\times 10^2$ $\uparrow$ \\
\midrule
4   & 147{,}474     & 0.78\%  & 96.10 & 60.60 \\
8   & 294{,}930     & 1.56\%  & 96.12 & 60.77 \\
16  & 589{,}842     & 3.12\%  & 96.10 & 60.55 \\
32  & 1{,}179{,}666 & 6.25\%  & 96.11 & 60.63 \\
64  & 2{,}359{,}314 & 12.50\% & 96.08 & 60.38 \\
128 & 4{,}718{,}610 & 25.00\% & 96.08 & 60.38 \\
\bottomrule
\end{tabular}
\end{table}

\begin{table}[!htbp]
\centering
\caption{Target-benchmark diagnostic rank sweep on DAVIS-all~\citep{davis17}. Parameter counts are reported across candidate rank hyperparameters \(k\), compared with the implemented full-rank quadratic probe at \(d=1024\) and \(L_{\mathrm{agg}}=18\) aggregated blocks. For $\pi^3$X-PMD, we report the best token-level Accuracy and Normalized Gain over all inter-frame blocks. This analysis is used only to diagnose whether the compact-rank behavior transfers to DAVIS-all, not for selecting \(k\).}
\label{tab:pmd_capacity}
\begin{tabular}{ccccc}
\toprule
Rank \(k\) & \(P_{\mathrm{PMD}}(k)\) & \(P_{\mathrm{PMD}}(k) / P_{\mathrm{full}}\) & Accuracy (\%) $\uparrow$ & NG $\times 10^2$ $\uparrow$ \\
\midrule
4   & 147{,}474     & 0.78\%  & 95.54 & 66.48 \\
8   & 294{,}930     & 1.56\%  & 95.57 & 66.70 \\
16  & 589{,}842     & 3.12\%  & 95.54 & 66.49 \\
32  & 1{,}179{,}666 & 6.25\%  & 95.60 & 66.95 \\
64  & 2{,}359{,}314 & 12.50\% & 95.48 & 66.01 \\
128 & 4{,}718{,}610 & 25.00\% & 95.36 & 65.09 \\
256 & 9{,}437{,}202 & 50.00\% & 95.30 & 64.66 \\
\bottomrule
\end{tabular}
\end{table}

Table~\ref{tab:pmd_capacity_val} reports the validation-based rank-selection analysis on the shared YouTube-VOS validation subset. We use this validation-only analysis to choose the default rank, avoiding any target-benchmark-specific rank selection. In this validation setting, \(k=4\) already recovers most of the attainable performance, while \(k=8\) gives the best Accuracy and Normalized Gain among the tested ranks. We therefore adopt \(k=8\) as the default rank for all main experiments.

\subsection{Shuffled-label Selectivity Control}
\label{app:selectivity}

To test selectivity to token--label correspondence, we permute motion labels within each frame group while preserving its label counts and retrain the same rank-8 $\pi^3$X probe on identical frozen features. Table~\ref{tab:probe_selectivity} reports pairwise compatibility accuracy and balanced accuracy on YouTube-VOS validation, averaged over the 18 inter-frame blocks. Balanced accuracy averages recall for the same-state and different-state pair classes. Parameter counts include all block-specific probes.

The shuffled-label control is close to the always-same baseline and chance balanced accuracy, whereas the correctly labeled probes retain gains above 11 percentage points at both capacities. This supports recoverability that depends on the correct token--motion-state correspondence under the restricted probe. It does not separate physical motion from correlated semantic or objectness information, nor establish an intrinsic low-dimensional subspace of raw tokens. The compactness analysis concerns the supervised probe-induced representation.

\subsection{Sample-Complexity Intuition}

The parameter counts above should be interpreted as a degree-of-freedom proxy rather than a formal generalization bound, especially because our video-derived token pairs are highly correlated. Under standard bounded-feature assumptions, capacity-based generalization bounds often scale with the effective degrees of freedom, up to constants, logarithmic factors, and assumptions on the data distribution~\citep{shalev2014understanding,mohri2018foundations,wainwright2019high}. A full-rank quadratic probe has capacity on the order of
\[
    P_{\mathrm{full}} = \mathcal{O}(L_{\mathrm{agg}}d^2),
\]
whereas the low-rank bilinear probe has capacity on the order of
\[
    P_{\mathrm{PMD}}(k) = \mathcal{O}(L_{\mathrm{agg}}kd).
\]

To put these complexity classes into perspective, let \(N\) denote the total number of sampled pairwise training examples. Our default pipeline samples \(2\times10^5\) token pairs per epoch for 16 epochs (Appendix~\ref{app:implementation}), yielding \(N \approx 3.2\times10^6\) highly correlated pairs derived from 160 videos. For the full-rank probe, \(P_{\mathrm{full}}\approx1.88\times10^7\), which is substantially larger than \(N\). This places the unconstrained quadratic probe in a high-dimensional regime that can be prone to overfitting, especially given the correlation among video-derived token pairs.

By contrast, at the default rank \(k=8\), PMD has \(P_{\mathrm{PMD}}(8)\approx2.95\times10^5\) trainable parameters, well below the number of sampled pairs. This reduction from the \(\mathcal{O}(d^2)\) regime to the \(\mathcal{O}(kd)\) regime provides a capacity-based explanation for why low-rank quadratic PMD can generalize efficiently from limited video diversity. This interpretation is consistent with the empirical rank ablations in Table~\ref{tab:pmd_capacity}.

\subsection{Comparison to GeoMotion Decoder Capacity}
\label{app:geomotion_capacity}

To complement the degree-of-freedom analysis above, we compare PMD with GeoMotion~\citep{geomotion}, a representative learned motion decoder built on top of a feed-forward 3D reconstruction model. This comparison is intended as a capacity and parameter-count comparison, not as an architecture-independent sample-complexity bound.

For transformer-style neural decoders with depth \(L\) and width \(d_{\mathrm{dec}}\), the dominant parameter term scales on the order of \(L d_{\mathrm{dec}}^2\). GeoMotion uses a motion decoder with \(d_{\mathrm{dec}}=1024\) and \(L=5\), whereas PMD uses one low-rank bilinear probe per aggregated block:
\begin{equation}
P_{\mathrm{PMD}}(k)
=
L_{\mathrm{agg}}(2kd+1)
=
18(2kd+1),
\end{equation}
with \(L_{\mathrm{agg}}=18\) and \(d=1024\).

We further compare against the actual trainable motion branch used by GeoMotion. This branch includes the confidence decoder, confidence head, and motion-aware decoder, with total parameter count
\begin{equation}
P_{\mathrm{GeoMotion}} = 80{,}320{,}196.
\end{equation}

\begin{table}[!htbp]
\centering
\caption{Comparison between PMD and the trainable GeoMotion~\citep{geomotion} motion head. The second column reports the concrete parameter ratio. The third column reports a simple degree-of-freedom proxy \(L_{\mathrm{agg}}kd /(L d_{\mathrm{dec}}^2)\), ignoring constant factors, with \(L_{\mathrm{agg}}=18\), \(d=d_{\mathrm{dec}}=1024\), and \(L=5\); it is not a formal sample-complexity bound.}
\label{tab:geomotion_capacity_ratio}
\begin{tabular}{ccc}
\toprule
PMD rank \(k\) & \(P_{\mathrm{GeoMotion}} / P_{\mathrm{PMD}}(k)\) & \(L_{\mathrm{agg}}kd /(L d_{\mathrm{dec}}^2)\,\downarrow\) \\
\midrule
4   & \num{5.45e2} & \num{1.41e-2} \\
8   & \num{2.72e2} & \num{2.81e-2} \\
16  & \num{1.36e2} & \num{5.63e-2} \\
32  & \num{6.81e1} & \num{1.13e-1} \\
64  & \num{3.40e1} & \num{2.25e-1} \\
128 & \num{1.70e1} & \num{4.50e-1} \\
256 & \num{8.51e0} & \num{9.00e-1} \\
\bottomrule
\end{tabular}
\end{table}

Thus, even at the largest candidate rank \(k=256\), PMD remains nearly one order of magnitude smaller than the trainable GeoMotion motion head, while in the practically relevant low-rank regime (\(k\le16\)) the concrete parameter gap remains above two orders of magnitude.

\section{Computational Efficiency and Inference Cost}
\label{app:efficiency}

To further highlight the lightweight and plug-and-play nature of PMD, we analyze its inference-time overhead. A major bottleneck in existing dynamic scene understanding pipelines is reliance on heavy neural decoders, iterative optimization, or auxiliary motion-estimation modules, which substantially increase memory footprint and computational cost. Table~\ref{tab:efficiency} summarizes the resulting trainable-parameter and inference-cost comparison.

\begin{table}[!htbp]
\centering
\caption{Inference efficiency comparison.}
\label{tab:efficiency}
\setlength{\tabcolsep}{3pt}
\renewcommand{\arraystretch}{1.15}

\renewcommand{\tabularxcolumn}[1]{m{#1}} 

\begin{tabularx}{\textwidth}{@{} l >{\raggedright\arraybackslash}X >{\raggedright\arraybackslash}X c c @{}}
\toprule
Method 
& \multicolumn{1}{c}{Backbone}
& \multicolumn{1}{c}{Trainable Module}
& \makecell[c]{Trainable Params $\downarrow$}
& \makecell[c]{Added FLOPs \\ / Frame $\downarrow$} \\
\midrule

\rowcolor{gray!10} 
\multicolumn{5}{l}{\textit{Large-scale supervised}} \\

SegAnyMo~\citep{seg_any_mo}
& DINOv2 + Depth Anything v2 + TAPIR
& Trajectory-based motion segmentation network
& \num{3.44} M
& \textgreater{}\num{2.08} G \\
\arrayrulecolor{gray!25}\midrule\arrayrulecolor{black}

Flow-SAM~\citep{flow_sam}
& SAM + RAFT
& FlowP-SAM decoder + FlowI-SAM decoder
& \num{9.83} M
& \textgreater{}\num{3.63e4} G \\
\arrayrulecolor{gray!25}\midrule\arrayrulecolor{black}

GeoMotion~\citep{geomotion}
& $\pi^3$ + RAFT
& Confidence decoder + confidence head + motion-aware decoder
& \num{80.3} M
& \textgreater{}\num{3.09e2} G \\

\midrule

\rowcolor{gray!10} 
\multicolumn{5}{l}{\textbf{\textit{Ours (PMD, $k=8$)}}} \\

PMD
& $\pi^3$
& Low-rank quadratic probe
& \textbf{\num{0.295} M}
& \textbf{\num{1.62} G} \\
\arrayrulecolor{gray!25}\midrule\arrayrulecolor{black}

PMD
& $\pi^3$X
& Low-rank quadratic probe
& \textbf{\num{0.295} M}
& \textbf{\num{1.62} G} \\

\bottomrule
\end{tabularx}
\end{table}

\paragraph{Measurement Protocol.}
In Tables~\ref{tab:unified_efficiency} and~\ref{tab:efficiency}, all added FLOPs are normalized per original video frame. For PMD, the reported Added FLOPs are computed directly from the current implementation and count only PMD-specific scoring and averaging beyond the frozen backbone. Concretely, they include probe scoring, pair averaging, and block averaging, while excluding backbone inference. For external methods, when full end-to-end inference costs cannot be cleanly isolated from the public implementation, we report conservative lower bounds under the official default inference configuration, denoted by \(>\). Trainable Params denote only the parameters optimized by the released training pipeline.

\paragraph{Why Some External Pipelines Are Expensive at Inference.}
A small trainable decoder does not necessarily imply low inference cost. In particular, Flow-SAM~\citep{flow_sam} combines optical-flow preprocessing with repeated prompt-conditioned decoding over a dense uniform grid of point prompts. Under the official default inference configuration, this leads to substantially higher per-frame FLOPs despite a relatively small trainable decoder. PMD, by contrast, evaluates lightweight pairwise probe scores against a fixed anchor token and then performs simple averaging, resulting in a much smaller added inference cost.

\paragraph{Discussion.}
As shown in Table~\ref{tab:efficiency}, PMD remains substantially smaller than competing learned motion-decoding pipelines under this protocol. Its core rank-8 probe uses only \(0.295\)M trainable parameters and adds \(1.62\)G FLOPs per original video frame, whereas heavier motion pipelines require either much larger trainable decoders or significantly more expensive inference-time procedures. This gap is especially pronounced relative to GeoMotion~\citep{geomotion}, which also builds on $\pi^3$ as its feed-forward 3D backbone but still optimizes an \(80.3\)M-parameter motion branch in its released frozen-backbone setting. It is also pronounced relative to Flow-SAM~\citep{flow_sam}, whose repeated grid-prompt decoding makes inference particularly expensive.

\paragraph{End-to-end Latency and Peak Memory.}
Table~\ref{tab:runtime_vram} reports batch-1 wall-clock latency and peak GPU memory on one RTX 5070 Ti with 16\,GB VRAM. Measurements include required preprocessing and backbone inference, use 10 warm-up frames and 100 measurement frames, and exclude model loading and result writes. Sequence-level runtimes are divided by 100 output frames. Each released pipeline otherwise retains its native settings, so these measurements compare complete implementations rather than equalized resolutions or module costs. Unlike the literature-quoted segmentation scores, these runtime measurements are obtained from the released pipelines on the same GPU.

\begin{table}[!htbp]
\centering
\caption{End-to-end inference measurements. SAM2 is included only in the corresponding refined pipelines; native SAM/SAM2 components are included for Flow-SAM and SegAnyMo.}
\label{tab:runtime_vram}
\small
\setlength{\tabcolsep}{4pt}
\renewcommand{\arraystretch}{1.12}
\begin{tabularx}{\textwidth}{@{}l >{\raggedright\arraybackslash}X cc@{}}
\toprule
Pipeline & Included inference scope & \makecell{Latency\\(ms/frame)} & \makecell{Peak VRAM\\(GiB)} \\
\midrule
$\pi^3$X backbone & Preprocessing and frozen-backbone forward & 480.6 & 4.47 \\
$\pi^3$X+coarse PMD & Above, plus probe, aggregation, upsampling, thresholding & 484.8 & 4.47 \\
$\pi^3$X+PMD+SAM2 & Above, plus sequence refinement & 952.8 & 4.47 \\
GeoMotion coarse~\citep{geomotion} & Released $\pi^3$ + RAFT + motion-decoder pipeline & 371.3 & 8.83 \\
GeoMotion+SAM2 & Above, plus released sequence refinement & 526.7 & 9.80 \\
Flow-SAM~\citep{flow_sam} & Released RAFT + FlowP/FlowI + SAM pipeline & 3{,}104.5 & 12.33 \\
SegAnyMo~\citep{seg_any_mo} & Released depth + appearance + tracking + SAM2 pipeline & 2{,}921.1 & 3.02 \\
\bottomrule
\end{tabularx}
\end{table}

Coarse PMD adds 4.2\,ms/frame (0.87\%) to $\pi^3$X inference, with no change in peak VRAM at the reported precision. Optional SAM2 adds 468.0\,ms/frame and dominates the additional latency. GeoMotion has lower total latency in both coarse and refined configurations despite its larger trainable motion branch, while SegAnyMo has lower measured peak VRAM. The efficiency claim therefore concerns PMD's small added readout cost and trainable capacity, not universal superiority in full-pipeline latency or memory.

\section{Broader Impacts}
\label{app:broader_impacts}

Our experiments show that motion-relevant information can be recovered from frozen 3D foundation models with low-capacity supervised probes and limited source-domain annotations.

\textbf{Positive impacts.}
By reducing the computational and data requirements for dynamic scene understanding, our approach can improve the accessibility and efficiency of research on 4D perception. This is particularly relevant for resource-constrained settings, where the lightweight extraction of motion cues from pretrained models may support applications in robotics, autonomous driving, and AR/VR systems. 

\textbf{Potential negative impacts.}
Like any robust motion segmentation and tracking technology, our method could be repurposed for unauthorized surveillance. The ability to track moving entities without requiring labeled target data could theoretically be applied to unconstrained CCTV footage for privacy-invading monitoring. We advocate that the deployment of such systems be accompanied by appropriate privacy-preserving safeguards.

\section{Additional Quantitative Results}
\label{app:additional_quantitative}

\subsection{Additional Layer-wise Probing Results}

\begin{figure}[!htbp]
    \centering
    \includegraphics[width=\linewidth]{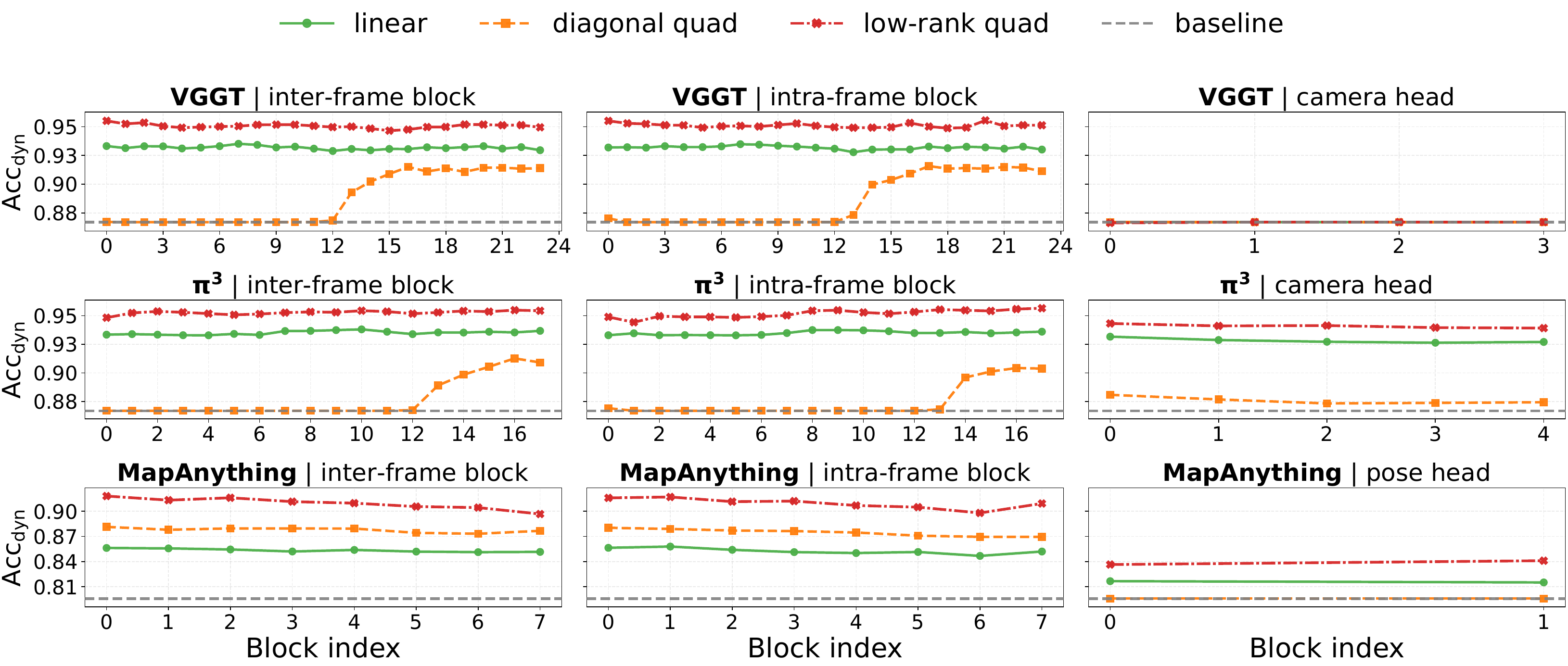}
    \caption{Additional layer-wise motion decodability on DAVIS-all~\citep{davis17}: VGGT~\citep{vggt}, $\pi^3$~\citep{pi3}, and MapAnything~\citep{mapanything}.}
    \label{fig:additional_layerwise_vggt_mapa}
\end{figure}

Figure~\ref{fig:additional_layerwise_vggt_mapa} complements the main layer-wise analysis by reporting results on VGGT~\citep{vggt}, $\pi^3$~\citep{pi3}, and MapAnything~\citep{mapanything}. The same trend holds beyond the three main backbones: low-rank quadratic probes consistently expose stronger motion decodability than linear or diagonal probes. VGGT's camera head collapses close to the all-static baseline, whereas $\pi^3$ shows a similar layer-wise pattern but retains decodable motion cues in its camera head, consistent with its highly similar yet more dynamic-scene-oriented training setup with datasets such as OmniWorld~\citep{omniworld}. However, for $\pi^3$, the motion decodability is substantially stronger in the geometric backbone blocks, including inter-frame and intra-frame blocks, than in the camera head. MapAnything similarly shows stronger motion decodability in geometric feature blocks than in its pose head.

\subsection{Additional Backbone Benchmarking}
\label{app:additional_backbone_benchmarking}

Table~\ref{tab:additional_pmd_backbones_compact} reports detailed PMD results across additional feed-forward 3D foundation model backbones, both with and without SAM2~\citep{sam2} refinement. These results demonstrate that the probing-based decoding framework is not specific to a single architecture. In these comparisons, $\pi^3$ and $\pi^3$X~\citep{pi3} combine strong coarse masks with small J-R gains from SAM2. Their larger J-M improvements show that refinement still affects mask quality substantially.

\begin{table}[!htbp]
\centering
\caption{
Additional PMD benchmarking of different feed-forward 3D foundation model backbones. Each benchmark cell reports J-Mean / J-Recall. \(\Delta\)J-R denotes the J-Recall gain from SAM2~\citep{sam2} refinement.
Avg. \(\Delta\)J-R and Std. \(\Delta\)J-R are computed over DAVIS2017~\citep{davis17}, SegTrackv2~\citep{segtrack}, and FBMS-59~\citep{fbms}; smaller gains indicate less change in the fraction of frames exceeding 0.5 Jaccard, not necessarily smaller changes in mask quality.
}
\label{tab:additional_pmd_backbones_compact}
\setlength{\tabcolsep}{4pt}
\renewcommand{\arraystretch}{1.12}
\begin{tabular}{l c c c c c c}
\toprule
Model 
& Refine 
& \makecell{DAVIS17 \\ (J-M / J-R) $\uparrow$}
& \makecell{SegTrackv2 \\ (J-M / J-R) $\uparrow$}
& \makecell{FBMS-59 \\ (J-M / J-R) $\uparrow$}
& \makecell{Avg. \\ \(\Delta\)J-R $\downarrow$}
& \makecell{Std. \\ \(\Delta\)J-R $\downarrow$} \\
\midrule

\multirow{2}{*}{$\pi^3$-PMD}
& w/o SAM2
& 72.7 / 89.9
& 72.9 / 92.9
& 72.6 / 83.4
& \multirow{2}{*}{2.59}
& \multirow{2}{*}{1.45} \\
& w/ SAM2
& 80.2 / 91.1
& 84.5 / 97.5
& 81.1 / 85.4
& & \\

\arrayrulecolor{gray!30}\midrule\arrayrulecolor{black}

\multirow{2}{*}{$\pi^3$X-PMD}
& w/o SAM2
& 73.7 / 91.0
& 72.9 / 94.0
& 73.5 / 88.8
& \multirow{2}{*}{2.37}
& \multirow{2}{*}{2.11} \\
& w/ SAM2
& 81.9 / 92.3
& 86.4 / 99.3
& 83.2 / 89.3
& & \\

\arrayrulecolor{gray!30}\midrule\arrayrulecolor{black}

\multirow{2}{*}{DA3-PMD}
& w/o SAM2
& 69.6 / 85.8
& 66.3 / 80.3
& 69.8 / 81.4
& \multirow{2}{*}{6.36}
& \multirow{2}{*}{2.96} \\
& w/ SAM2
& 77.7 / 88.2
& 75.7 / 89.8
& 82.3 / 88.6
& & \\

\arrayrulecolor{gray!30}\midrule\arrayrulecolor{black}

\multirow{2}{*}{DUSt3R-PMD}
& w/o SAM2
& 62.2 / 70.0
& 58.6 / 77.3
& 64.0 / 74.5
& \multirow{2}{*}{3.74}
& \multirow{2}{*}{2.50} \\
& w/ SAM2
& 66.5 / 70.6
& 66.0 / 84.0
& 74.0 / 78.3
& & \\

\arrayrulecolor{gray!30}\midrule\arrayrulecolor{black}

\multirow{2}{*}{VGGT-PMD}
& w/o SAM2
& 72.6 / 89.9
& 70.8 / 84.7
& 73.5 / 85.9
& \multirow{2}{*}{6.35}
& \multirow{2}{*}{3.92} \\
& w/ SAM2
& 80.8 / 92.3
& 83.6 / 96.4
& 82.9 / 90.8
& & \\

\arrayrulecolor{gray!30}\midrule\arrayrulecolor{black}

\multirow{2}{*}{MapA-PMD}
& w/o SAM2
& 69.2 / 86.0
& 62.3 / 75.7
& 69.1 / 79.9
& \multirow{2}{*}{3.95}
& \multirow{2}{*}{6.11} \\
& w/ SAM2
& 77.3 / 86.6
& 75.5 / 88.2
& 75.8 / 78.5
& & \\

\bottomrule
\end{tabular}
\end{table}

\subsection{Backbone Disparity and Source-validation Decodability}
\label{app:backbone_disparity}

To compare recoverability across backbones before mask extraction, we evaluate rank-8 quadratic probes on the same 80-video YouTube-VOS validation split~\citep{youtubevos} and average over all blocks in each fixed PMD-selected group. We distinguish pairwise normalized gain from the unary dynamic-token NG defined in Section~\ref{sec:setup}:
\[
\mathrm{NG}_{\mathrm{pair}}
=\frac{\mathrm{Acc}_{\mathrm{pair}}-\mathrm{Baseline}_{\mathrm{same}}}
{1-\mathrm{Baseline}_{\mathrm{same}}},
\qquad
\mathrm{Baseline}_{\mathrm{same}}
=\frac{1}{|\mathcal{P}|}\sum_{(i,j)\in\mathcal{P}}\mathbf{1}[z_{ij}=1].
\]
Here, $\mathrm{Acc}_{\mathrm{pair}}$ is the accuracy of thresholding pairwise compatibility at $0.5$, and $\mathrm{Baseline}_{\mathrm{same}}$ always predicts that a pair shares the same motion state. These measurements use frozen tokens before thresholding score maps or applying SAM2. Target J-M is the mean coarse-mask J-M across DAVIS17, SegTrackv2, and FBMS-59.

\begin{table}[!htbp]
\centering
\caption{Source-validation pairwise decodability and target coarse-mask performance. Source NG is $\mathrm{NG}_{\mathrm{pair}}\times10^2$ (mean $\pm$ standard deviation across blocks in each selected group); target J-M is averaged over DAVIS17, SegTrackv2, and FBMS-59.}
\label{tab:backbone_ng}
\footnotesize
\setlength{\tabcolsep}{4pt}
\renewcommand{\arraystretch}{1.16}
\renewcommand{\tabularxcolumn}[1]{m{#1}}
\begin{tabularx}{\textwidth}{@{}l >{\raggedright\arraybackslash}X r@{\,$\pm$\,}l r@{}}
\toprule
\textbf{Backbone} & \textbf{Architecture} & \multicolumn{2}{c}{\makecell{\textbf{Source-val. NG}\\mean $\pm$ std.}} & \makecell{\textbf{Target J-M}\\($\uparrow$)} \\
\midrule
\rowcolor{gray!10}\multicolumn{5}{l}{\textit{Multi-view backbones}\quad (probe group: inter-frame)} \\
$\pi^3$X~\citep{pi3} & DINOv2-L + 36 alternating frame/global blocks & 72.46 & 2.06 & 73.35 \\
\addlinespace[2pt]
$\pi^3$~\citep{pi3} & DINOv2-L + 36 alternating frame/global blocks & 76.76 & 0.78 & 72.77 \\
\addlinespace[2pt]
VGGT~\citep{vggt} & DINOv2-L + 24 paired frame/global stages & 73.60 & 4.18 & 72.31 \\
\addlinespace[2pt]
DA3~\citep{da3} & 40-block ViT-G; local/global alternation from block 13 & 61.87 & 6.60 & 68.58 \\
\addlinespace[2pt]
MapAnything~\citep{mapanything} & DINOv2-G + separate 16-block UniCeption stack & 67.20 & 2.10 & 66.86 \\
\midrule
\rowcolor{gray!10}\multicolumn{5}{l}{\textit{Pairwise backbone}\quad (probe group: decoder/head view 1)} \\
DUSt3R~\citep{dust3r} & 24-block CroCo encoder + two 12-block reciprocal decoders & 33.73 & 5.89 & 61.59 \\
\bottomrule
\end{tabularx}
\end{table}

Relative to DA3, $\pi^3$X has 10.59 points higher mean source-validation NG, 4.54 points lower across-block standard deviation, and 4.77 points higher mean target J-M (Table~\ref{tab:backbone_ng}). The same restricted probe therefore recovers stronger and more uniform motion-state information from the selected $\pi^3$X tokens. On DAVIS17, the coarse $\pi^3$X--DA3 gap is 4.1 J-M, close to the 4.2-point gap after SAM2; the disparity is already present before refinement. Excluding DUSt3R reduces the cross-backbone standard deviation of three-benchmark mean J-M from 4.14 to 2.57, computed from unrounded benchmark means. This identifies DUSt3R as a relative outlier while leaving variation among the multi-view backbones.

\begin{table}[!htbp]
\centering
\caption{Descriptive grouping by broad architectural design regime. Source-validation NG and target J-M are averaged over the backbones in each row.}
\label{tab:design_regimes}
\footnotesize
\setlength{\tabcolsep}{3pt}
\renewcommand{\arraystretch}{1.12}
\begin{tabular}{@{}>{\raggedright\arraybackslash}p{0.45\linewidth}>{\raggedright\arraybackslash}p{0.20\linewidth}cc@{}}
\toprule
Broad design regime & Backbones & \makecell{Mean source\\NG $\times10^2$} & \makecell{Mean target\\J-M} \\
\midrule
Separate ViT-L encoder + alternating frame/global multi-view stack & $\pi^3$X, $\pi^3$, VGGT & 74.27 & 72.81 \\
ViT-G-scale local/global multi-view stack + dense geometry heads & DA3, MapAnything & 64.54 & 67.72 \\
Pairwise asymmetric encoder--decoder with reciprocal cross-attention & DUSt3R & 33.73 & 61.59 \\
\bottomrule
\end{tabular}
\end{table}

Table~\ref{tab:design_regimes} groups broad design regimes rather than strict architecture families. In particular, DA3 and MapAnything differ in token flow, supported input modalities, and prediction heads despite sharing a ViT-G-scale local/global geometry-decoding pattern. Source-validation pairwise NG is positively associated with coarse target J-M, including within the five multi-view backbones (Table~\ref{tab:backbone_correlations}).

\begin{table}[!htbp]
\centering
\caption{Descriptive associations between source-validation pairwise NG and mean target coarse-mask J-M.}
\label{tab:backbone_correlations}
\small
\setlength{\tabcolsep}{8pt}
\begin{tabular}{@{}lcc@{}}
\toprule
Comparison & Spearman $\rho$ & Pearson $r$ \\
\midrule
All six backbones & 0.77 & 0.93 \\
Five multi-view backbones & 0.60 & 0.80 \\
Three design-regime means & 1.00 & 0.97 \\
\bottomrule
\end{tabular}
\end{table}

High mean source-validation NG and low across-block dispersion may help identify promising candidates across broad design regimes without target-benchmark measurements. They do not determine the exact ordering of closely related backbones: $\pi^3$ has higher source-validation NG than $\pi^3$X, whereas $\pi^3$X has 0.58 points higher target J-M. These small-sample associations are descriptive; architecture, pretraining data, token flow, and prediction heads vary together, so the analysis does not isolate a causal architectural driver of motion recoverability.

\subsection{Additional Feed-forward Geometry-based Comparisons}
\label{app:ff_geometry_comparison}

Table~\ref{tab:ff_geometry_motion_comparison_v2} complements the main benchmark by isolating feed-forward geometry-based motion segmentation methods and reporting their performance with and without SAM2~\citep{sam2} refinement. This comparison is useful because these methods operate in a similar regime: they derive motion cues from feed-forward 3D foundation models. We therefore separate the quality of the coarse motion cues from the effect of SAM2 refinement by reporting both J-Mean and J-Recall before and after refinement, as well as the J-Recall gain \(\Delta\)J-R. For methods with high coarse J-R, a small \(\Delta\)J-R means that refinement moves relatively few additional frames above the 0.5 Jaccard threshold. It does not measure pixel recall or imply negligible improvement in mask extent or boundaries.

\begin{table}[!htbp]
\centering
\caption{Feed-forward geometry-based motion segmentation comparison on DAVIS2017~\citep{davis17}. 
We report J-Mean (J-M), J-Recall (J-R), and the J-Recall gain from SAM2 refinement. $\Delta$J-R measures changes in the fraction of frames above 0.5 Jaccard; J-M separately captures changes in average mask overlap. Note that PMD introduces substantially fewer additional parameters and lower FLOP overhead than GeoMotion; see Table~\ref{tab:efficiency}.}
\label{tab:ff_geometry_motion_comparison_v2}
\setlength{\tabcolsep}{6pt}
\begin{tabular}{l c cc cc c}
\toprule
\multirow{2}{*}{Method} 
& \multirow{2}{*}{Flow} 
& \multicolumn{2}{c}{w/o SAM2} 
& \multicolumn{2}{c}{w/ SAM2}
& \multirow{2}{*}{$\Delta$J-R $\downarrow$} \\
\cmidrule(lr){3-4} \cmidrule(lr){5-6}
& 
& J-M $\uparrow$ 
& J-R $\uparrow$ 
& J-M $\uparrow$ 
& J-R $\uparrow$
& \\
\midrule
Easi3R\textsubscript{dust3r}~\citep{easi3r}   
& \xmark 
& 46.9 
& 50.5 
& 60.1 
& 65.3
& 14.8 \\

Easi3R\textsubscript{monst3r}~\citep{easi3r}  
& \xmark 
& 54.8 
& 66.2 
& 67.9 
& 76.1
& 9.9 \\

MonST3R~\citep{monst3r}                       
& \cmark 
& 38.1 
& 36.1 
& 56.4 
& 59.6
& 23.5 \\

DAS3R~\citep{das3r}                                        
& \xmark 
& 44.5 
& 44.0 
& 57.4 
& 61.3
& 17.3 \\

VGGT4D~\citep{vggt4d}                         
& \xmark 
& 56.5 
& 65.6 
& --    
& --    
& -- \\

GeoMotion~\citep{geomotion}                   
& \cmark 
& \textbf{75.4} 
& 87.2 
& 81.1 
& \textbf{92.6}
& 5.4 \\

\midrule
\rowcolor{gray!10}
\multicolumn{7}{l}{\textit{Ours}} \\

$\pi^3$-PMD (ours)                                 
& \xmark 
& 72.7 
& 89.9 
& 80.2 
& 91.1
& \textbf{1.2} \\

$\pi^3$X-PMD (ours)                                 
& \xmark 
& 73.7 
& \textbf{91.0} 
& \textbf{81.9} 
& 92.3
& 1.3 \\
\bottomrule
\end{tabular}
\end{table}

\subsection{Additional Scene-Diversity Starvation Results}

\begin{figure}[!htbp]
    \centering
    \includegraphics[width=\linewidth]{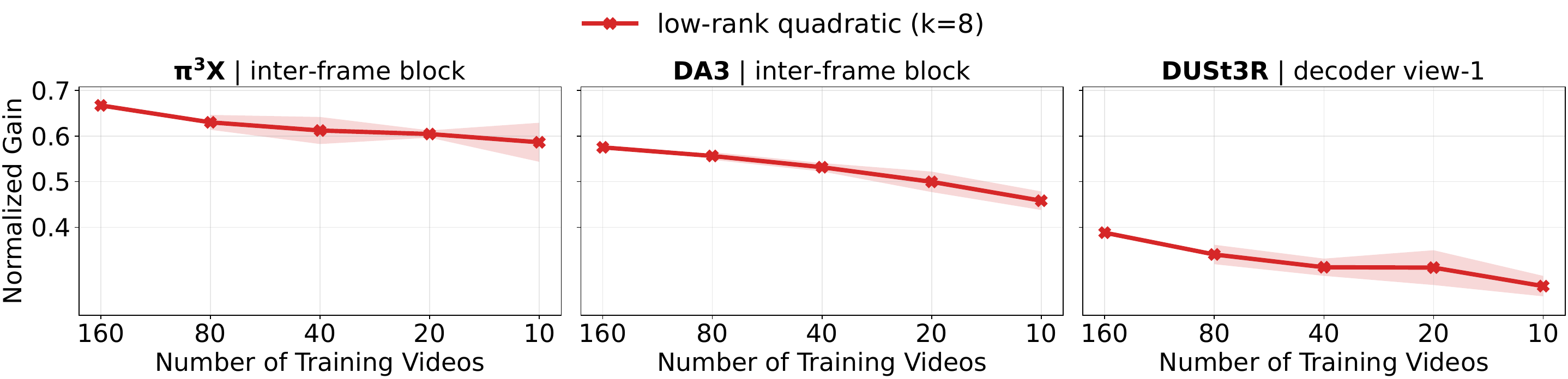}
    \caption{
    Additional scene-diversity starvation analysis on DAVIS-all~\citep{davis17}. We report token-level Normalized Gain for the default low-rank quadratic probe (\(k=8\)) across $\pi^3$X~\citep{pi3}, DA3~\citep{da3}, and DUSt3R~\citep{dust3r} while reducing the number of YouTube-VOS training videos. Solid lines denote the mean over available random subset seeds, and shaded bands denote one standard deviation when multiple seeds are available. Performance degrades gradually as training diversity is reduced, while the uncertainty bands remain moderate overall and widen mainly in the most data-starved regimes.
    }
    \label{fig:app_data_starvation_ng_k8}
\end{figure}

Figure~\ref{fig:app_data_starvation_ng_k8} provides additional token-level scene-diversity starvation results for the default low-rank quadratic probe with \(k=8\). Across $\pi^3$X~\citep{pi3}, DA3~\citep{da3}, and DUSt3R~\citep{dust3r}, normalized gain decreases gradually as the number of training videos is reduced, further supporting that the decoded motion cues are compact and can be recovered with limited scene diversity. Among the three backbones, $\pi^3$X achieves the highest normalized gain and degrades most slowly under reduced training data. This observation is consistent with the quantitative evaluation in Table~\ref{tab:sota_full}, where the $\pi^3$X-PMD variant achieves the strongest downstream motion segmentation performance. To assess sensitivity to subset choice, we additionally report the mean and one-standard-deviation band over three random subset seeds whenever multiple seeds are available. The uncertainty remains modest overall and increases mainly in the most data-starved settings, indicating graceful degradation rather than unstable collapse.

\subsection{Extended Benchmark Results}
\label{app:full_benchmark}
The following table retains the full set of benchmark scores, including DAVIS16 and DAVIS-all. Refinement use and target-domain training are marked explicitly; the grouped comparison in the main text provides additional context on supervision and inference-time priors.

\begin{table}[!htbp]
\centering
\small
\caption{Extended motion segmentation benchmarking. All non-PMD scores are quoted from the cited papers; only PMD rows are evaluated with our evaluator. These complete-pipeline results have differing supervision and inference priors, separated in Table~\ref{tab:sota_full}. Trained on only 160 YouTube-VOS 2018~\citep{youtubevos} videos, PMD transfers across DAVIS~\citep{davis16,davis17}, SegTrackv2~\citep{segtrack}, and FBMS-59~\citep{fbms}; its strongest multi-view pretrained variants approach supervised and self-supervised methods. \textcolor{gray}{Gray entries} denote in-domain training, fine-tuning, or test-time optimization on the target benchmark. Methods using SAM/SAM2~\citep{sam,sam2}, either for external refinement or as a native pipeline component, are marked with $^\dagger$.}
\label{tab:full_benchmark}
\renewcommand{\arraystretch}{1.1} 
\setlength{\tabcolsep}{3pt}
\renewcommand{\tabularxcolumn}[1]{m{#1}}
\begin{tabularx}{\textwidth}{@{} X c c c c c @{}}
\toprule
\makecell[l]{Method} 
& \makecell[c]{DAVIS16 \\ (J-M $\uparrow$)} 
& \makecell[c]{DAVIS17 \\ (J-M $\uparrow$)} 
& \makecell[c]{DAVIS-all \\ (J-M $\uparrow$)} 
& \makecell[c]{SegTrackv2 \\ (J-M $\uparrow$)} 
& \makecell[c]{FBMS-59 \\ (J-M $\uparrow$)} \\
\midrule

\rowcolor{gray!10} \multicolumn{6}{l}{\textit{Unsupervised / Self-supervised}} \\
CIS~\citep{cis}        & \textcolor{gray}{71.5} & -- & -- & \textcolor{gray}{62.0} & \textcolor{gray}{63.5} \\
OCLR~\citep{oclr}      & \textcolor{gray}{80.9} & -- & -- & \textcolor{gray}{72.3} & \textcolor{gray}{69.8} \\
DyStaB~\citep{dystab}  & \textcolor{gray}{82.8} & -- & -- & \textcolor{gray}{72.7} & \textcolor{gray}{68.6} \\
RCF~\citep{rcf}        & \textcolor{gray}{80.9} & -- & -- & \textcolor{gray}{76.7} & \textcolor{gray}{69.9} \\
OCMOS \citep{ocmos}    & \textcolor{gray}{81.1} & -- & -- & \textcolor{gray}{76.6} & \textcolor{gray}{81.9} \\
EM-OFS~\citep{em_ofs}  & 75.2 & -- & -- & 65.5 & 71.3 \\
FlowFeat~\citep{flowfeat} & -- & 62.3 & -- & -- & -- \\
\midrule

\rowcolor{gray!10} \multicolumn{6}{l}{\textit{Training-free / Heuristic}} \\
ARP~\citep{arp}                              & 76.2 & -- & -- & 57.2 & 59.8 \\
VGGT4D \citep{vggt4d}                        & 62.1 & 56.5 & 50.75 & -- & -- \\
Easi3R\textsubscript{dust3r}$^\dagger$ \citep{easi3r} & 67.9 & 60.1 & 54.7 & -- & -- \\
Easi3R\textsubscript{monst3r}$^\dagger$ \citep{easi3r} & 70.7 & 67.9 & 63.1 & -- & -- \\
\midrule

\rowcolor{gray!10} \multicolumn{6}{l}{\textit{Supervised}} \\
MATNet~\citep{matnet}              & \textcolor{gray}{82.4} & -- & -- & \textcolor{gray}{50.4} & \textcolor{gray}{76.1} \\
COSNet~\citep{cosnet}              & \textcolor{gray}{80.5} & -- & -- & 49.7 & 75.6 \\
AMC-Net~\citep{amc_net}& \textcolor{gray}{84.5} & -- & -- & -- & 76.5 \\
Flow-SAM$^\dagger$~\citep{flow_sam} 
& \textcolor{gray}{86.2} 
& \textcolor{gray}{71.6} 
& -- 
& 84.2 
& \textbf{88.7} \\
SegAnyMo$^\dagger$~\citep{seg_any_mo}        & \textbf{90.6} & -- & -- & 76.3 & 78.3 \\
GeoMotion$^\dagger$~\citep{geomotion}        & 84.5 & 81.1 & 74.8 & 77.3 & 72.5 \\
\midrule

\rowcolor{gray!10} \multicolumn{6}{l}{\textbf{\textit{Ours (data-efficient probing)}}} \\
$\pi^3$X-PMD$^\dagger$ & 81.4 & \textbf{81.9} & \textbf{80.2} & \textbf{86.4} & 83.2 \\
$\pi^3$-PMD$^\dagger$ & 79.5 & 80.2 & 79.6 & 84.5 & 81.1 \\
DA3-PMD$^\dagger$ & 75.9 & 77.7 & 75.7 & 75.7 & 82.3 \\
DUSt3R-PMD$^\dagger$ & 61.5 & 66.5 & 63.8 & 66.0 & 74.0 \\
\bottomrule
\end{tabularx}
\end{table}

\section{Additional Qualitative Results}
\label{app:qualitative_results}

In this section, we provide extended qualitative visualizations to complement the findings in the main paper. We focus on: 
(1) the effect of network depth and scene-diversity starvation on motion--semantic entanglement, 
and (2) an analysis of failure cases and current limitations.

\subsection{Visualization of the Effect of Network Depth and Scene-Diversity Starvation}
\begin{figure}[!htbp]
    \centering
    \begin{subfigure}[b]{0.17\linewidth}
        \centering
        \includegraphics[width=\linewidth]{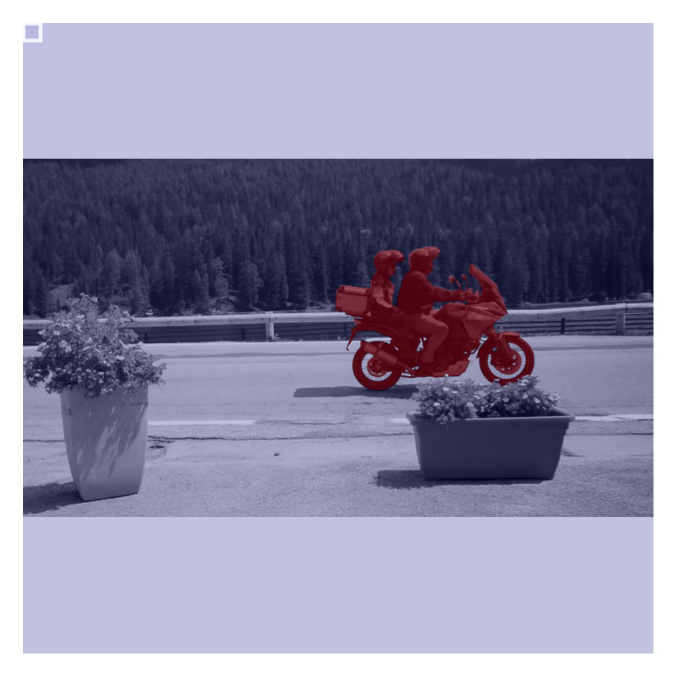}
        \caption{Ground Truth}
        \label{fig:gt_layer}
    \end{subfigure}
    \hfill
    \begin{subfigure}[b]{0.36\linewidth}
        \centering
        \includegraphics[width=\linewidth]{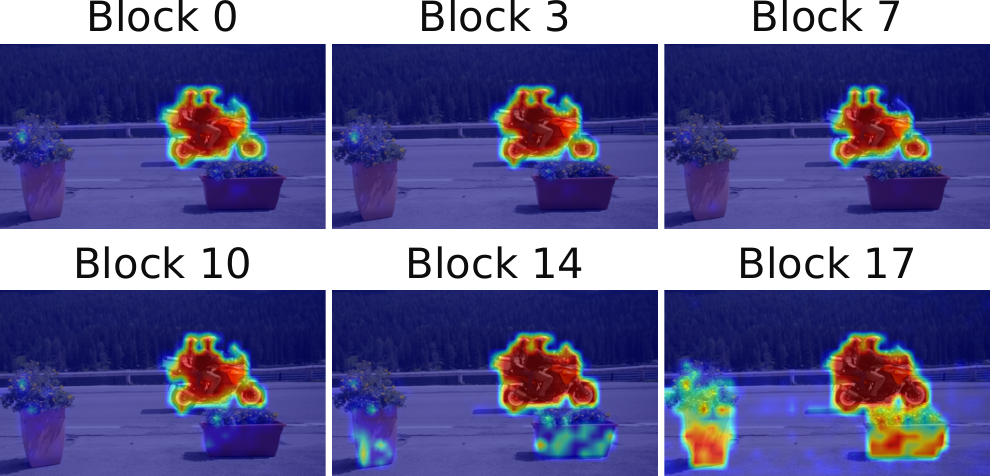}
        \caption{Trained with 160 videos}
    \end{subfigure}
    \hfill
    \begin{subfigure}[b]{0.36\linewidth}
        \centering
        \includegraphics[width=\linewidth]{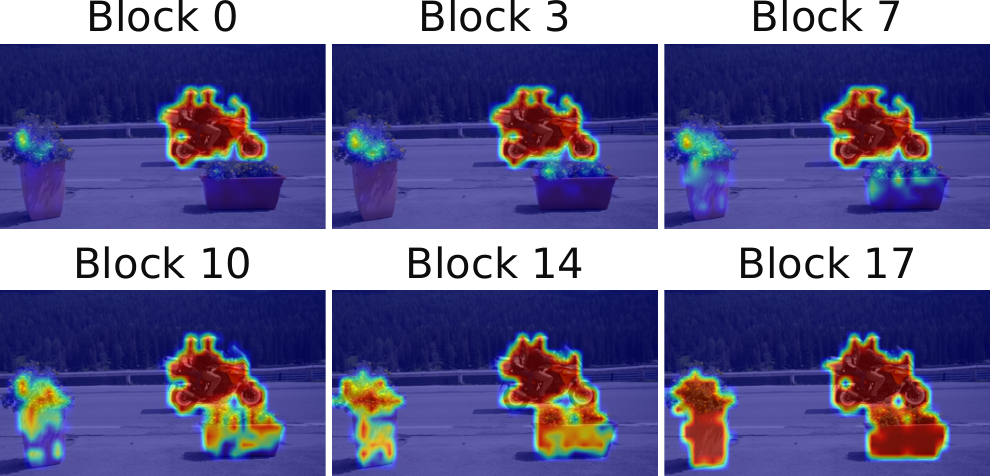}
        \caption{Trained with 10 videos}
    \end{subfigure}
    \caption{Scene-diversity starvation and layer-wise semantic leakage on $\pi^3$X~\citep{pi3}. Visualization of PMD dynamic score maps across network depth under varying amounts of training data. While deeper representations entangle motion with semantics, reducing the training subset to only 10 videos exacerbates this leakage across all layers. The starved probe incorrectly highlights static foreground objects, whereas the 160-video model maintains a focus on actual motion, particularly in its shallow and intermediate layers.}
    \label{fig:app_layer_evolution}
\end{figure}

Figure~\ref{fig:app_layer_evolution} provides visual evidence of how the decoded motion cues and motion--semantic entanglement are affected by both network depth and training-set size, using the $\pi^3$X backbone~\citep{pi3} as a representative example.

We compare the ground-truth motion masks (shown in Figure~\ref{fig:gt_layer}) against $\pi^3$X-PMD dynamic score maps extracted across different network layers, contrasting our default 160-video training subset with a severely starved 10-video subset. As shown, the 160-video model highlights the moving target in its shallow and intermediate layers. However, deeper layers begin to entangle motion with semantic category information, consistent with the increased effective rank required for decoding in deeper blocks (Figure~\ref{fig:metric_rank}). Under extreme data scarcity (10 videos), this entanglement becomes more pronounced across layers: the starved probe assigns high dynamic scores to static foreground instances rather than strictly isolating actual moving objects.

\subsection{Failure Cases and Limitation Analysis}
\label{app:limitations}

\begin{figure}[!htbp]
    \centering
    \begin{subfigure}{0.32\linewidth}
        \centering
        \includegraphics[width=\linewidth]{Graphics/Pi3X/breakdance/00032.jpg}
        \caption{Input frame}
    \end{subfigure}
    \hfill
    \begin{subfigure}{0.32\linewidth}
        \centering
        \includegraphics[width=\linewidth]{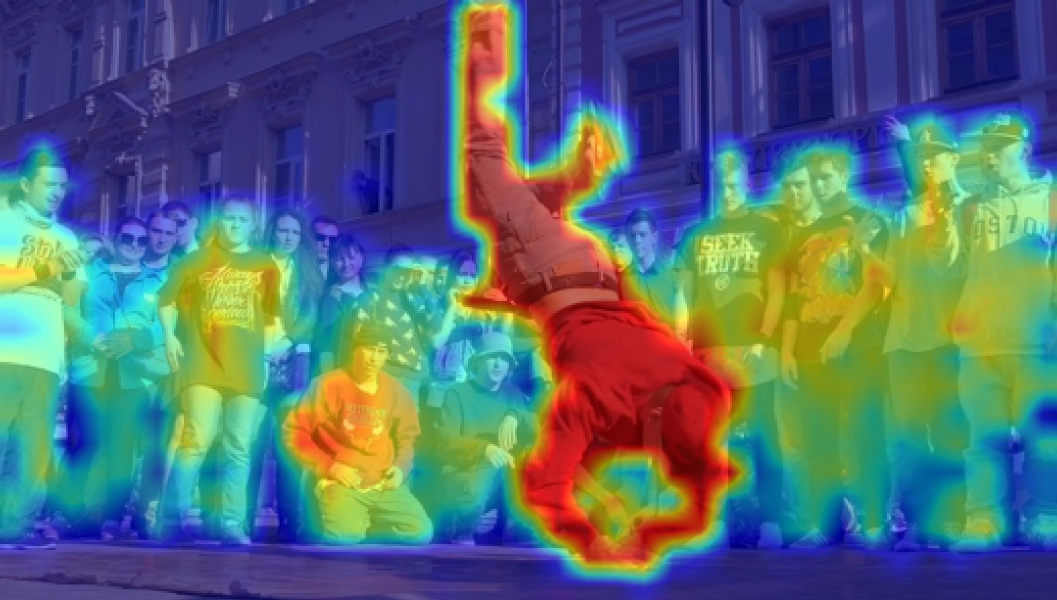}
        \caption{PMD dynamic score map}
    \end{subfigure}
    \hfill
    \begin{subfigure}{0.32\linewidth}
        \centering
        \includegraphics[width=\linewidth]{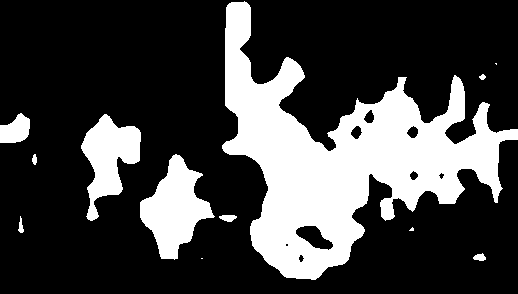}
        \caption{Binary mask}
    \end{subfigure}
    \caption{\textbf{Limitation: Fine-grained Motion vs. Binary Detection.} PMD highlights the breakdancer with the most prominent motion and assigns moderately elevated scores to nearby people with subtler in-place motion relative to the static background. However, the pipeline ultimately produces a binary static--dynamic mask rather than distinguishing different motion types or motion magnitudes.}
    \label{fig:app_limitation_fine_motion}
\end{figure}

Figure~\ref{fig:app_limitation_leakage} presents a semantic-leakage example discussed in Section~\ref{sec:limitations}. Here, Figure~\ref{fig:app_limitation_fine_motion} illustrates the complementary limitation of binary motion detection. These cases reflect the binary static--dynamic formulation, possible motion--semantic entanglement in frozen representations, and object-centric source supervision.

\paragraph{Coarse Binary Detection vs. Fine-Grained Motion.} 
PMD is designed to identify and segment dynamically moving regions. As illustrated in Figure~\ref{fig:app_limitation_fine_motion}, it does not currently distinguish fine-grained motion patterns, such as in-place articulation, localized deformation, or different motion magnitudes, or decode explicit physical kinematics, such as 3D displacement or velocity vectors.

\paragraph{Semantic Leakage.} 
The probe sometimes struggles to fully decouple motion from semantics. As shown in Figure~\ref{fig:app_limitation_leakage}, in scenes containing multiple instances of the same object category, static instances may receive erroneously elevated dynamic scores when another instance of that category is moving. Notably, this semantic leakage can occur even for categories absent from the 160-video training subset, consistent with entanglement of motion and category-level cues. Object-centric YouTube-VOS supervision may also contribute; the present experiments do not isolate these sources of bias or establish pure geometric dynamicity.

\paragraph{Evaluation Scope.}
The present evaluation covers frozen feed-forward 3D reconstruction backbones. Controlled comparisons with video segmentation and tracking foundation models under matched supervision and prompting would address differences in pretraining objectives, architectures, and decoding pipelines. Whether the observed recoverability is specific to 3D pretraining remains open. Training beyond 160 YouTube-VOS videos has not been evaluated, leaving performance saturation, an optimal data budget, and the benefits of additional source diversity unresolved.

\section{Existing Assets, Licenses, and Terms of Use}
\label{app:asset_licenses}
In this section, we provide detailed information regarding the existing assets used in our research to document compliance with the NeurIPS guidelines on data and model usage. Table~\ref{tab:asset_licenses_models} outlines the licenses and specific terms of use for all pretrained models and tools integrated into our pipeline. Similarly, Table~\ref{tab:asset_licenses_datasets} summarizes the licensing details for the datasets and benchmarks used for training and evaluation. We obtained these assets from their official sources and used them for non-commercial research evaluation, following the corresponding licenses and terms of use to the best of our understanding.

\begingroup
\small
\setlength{\tabcolsep}{4pt}
\renewcommand{\arraystretch}{1.10}
\begin{longtable}{@{}>{\raggedright\arraybackslash}p{0.22\textwidth} >{\raggedright\arraybackslash}p{0.18\textwidth} >{\raggedright\arraybackslash}p{\dimexpr0.60\textwidth-16pt\relax}@{}}
\caption{Existing pretrained models and tools used in this work. We obtain all models and tools from their official sources and follow their respective licenses and terms of use.}
\label{tab:asset_licenses_models}\\
\toprule
Asset & Use in this work & License / terms \\
\midrule
\endfirsthead
\multicolumn{3}{l}{\small\tablename~\thetable\ (continued)}\\
\toprule
Asset & Use in this work & License / terms \\
\midrule
\endhead
\midrule
\multicolumn{3}{r}{\small Continued on next page}\\
\endfoot
\bottomrule
\endlastfoot

SAM2~\citep{sam2}
& Optional mask refinement
& SAM2 code and model weights are released under Apache-2.0; the SAM2 evaluation code is released under BSD-3-Clause. We use SAM2 only as an optional refinement module and follow the official license terms. \\
\arrayrulecolor{gray!30}\midrule\arrayrulecolor{black}

$\pi^3$/$\pi^3$X~\citep{pi3}
& Frozen backbone
& Code is released under BSD-3-Clause; $\pi^3$ and $\pi^3$X model weights are released under CC BY-NC 4.0 and are restricted to non-commercial research and educational use. \\
\arrayrulecolor{gray!30}\midrule\arrayrulecolor{black}

Depth Anything 3~\citep{da3}
& Frozen backbone
& Code is released under Apache-2.0. The DA3-GIANT-1.1 model weights used in this work are released under CC BY-NC 4.0 and are restricted to non-commercial use. \\
\arrayrulecolor{gray!30}\midrule\arrayrulecolor{black}

DUSt3R~\citep{dust3r}
& Frozen backbone
& Code and released materials are distributed under CC BY-NC-SA 4.0, with non-commercial use only. We use DUSt3R only for research evaluation. \\
\arrayrulecolor{gray!30}\midrule\arrayrulecolor{black}

VGGT~\citep{vggt}
& Frozen backbone
& The VGGT repository is distributed under the custom VGGT License. According to the official repository, only the newer VGGT-1B-Commercial checkpoint is licensed for commercial usage, whereas the original VGGT-1B checkpoint remains non-commercial. Our experiments use the original VGGT-1B checkpoint only for non-commercial research evaluation and follow the corresponding license terms. \\
\arrayrulecolor{gray!30}\midrule\arrayrulecolor{black}

MapAnything~\citep{mapanything}
& Frozen backbone
& Code is released under Apache-2.0. The official pretrained model variants have separate licenses: \texttt{facebook/map-anything} is CC BY-NC 4.0, while \texttt{facebook/map-anything-apache} is Apache-2.0. We follow the license of the specific checkpoint used in each experiment. \\

\end{longtable}
\endgroup

\begingroup
\small
\setlength{\tabcolsep}{4pt}
\renewcommand{\arraystretch}{1.10}
\begin{longtable}{@{}>{\raggedright\arraybackslash}p{0.22\textwidth} >{\raggedright\arraybackslash}p{0.18\textwidth} >{\raggedright\arraybackslash}p{\dimexpr0.60\textwidth-16pt\relax}@{}}
\caption{Existing datasets and benchmarks used in this work. We obtain all datasets from their official sources and follow their respective licenses and terms of use.}
\label{tab:asset_licenses_datasets}\\
\toprule
Asset & Use in this work & License / terms \\
\midrule
\endfirsthead
\multicolumn{3}{l}{\small\tablename~\thetable\ (continued)}\\
\toprule
Asset & Use in this work & License / terms \\
\midrule
\endhead
\midrule
\multicolumn{3}{r}{\small Continued on next page}\\
\endfoot
\bottomrule
\endlastfoot

YouTube-VOS 2018~\citep{youtubevos}
& Probe training and validation source
& Annotations are licensed under CC BY 4.0; data are released for non-commercial research purposes only. We use the dataset only for research and obtain it from the official source. \\
\arrayrulecolor{gray!30}\midrule\arrayrulecolor{black}

DAVIS 2016/2017~\citep{davis16,davis17}
& Evaluation benchmark
& DAVIS 2017 challenge annotations are licensed under CC BY 4.0; the official DAVIS 2017 evaluation code is released under BSD-3-Clause. We use the dataset only for benchmark evaluation and cite the official DAVIS papers. \\
\arrayrulecolor{gray!30}\midrule\arrayrulecolor{black}

SegTrackv2~\citep{segtrack}
& Evaluation benchmark
& The official dataset page does not explicitly state a standard license. We use SegTrackv2 only for non-commercial research evaluation, obtain it from the official project page, and cite the original dataset paper. \\
\arrayrulecolor{gray!30}\midrule\arrayrulecolor{black}

FBMS-59~\citep{fbms}
& Evaluation benchmark
& Custom research-only terms: the dataset is provided only for research purposes and without warranty; commercial use is prohibited. We use FBMS-59 only for research evaluation and cite the corresponding dataset paper. \\

\end{longtable}
\endgroup

\section*{NeurIPS Paper Checklist}

\begin{enumerate}

\item {\bf Claims}
    \item[] Question: Do the main claims made in the abstract and introduction accurately reflect the paper's contributions and scope?
    \item[] Answer: \answerYes{}
    \item[] Justification: The abstract and introduction state the main claims: motion-relevant information can be decoded from frozen feed-forward 3D foundation models using low-capacity probes, and PMD achieves data-efficient zero-shot transfer to standard motion segmentation benchmarks. These claims are supported by the layer-wise probing, benchmark comparisons, data-starvation analysis, and compactness analysis in Sections~\ref{sec:origin}--\ref{sec:mechanism}; limitations are discussed in the Limitations section.
    \item[] Guidelines:
    \begin{itemize}
        \item The answer \answerNA{} means that the abstract and introduction do not include the claims made in the paper.
        \item The abstract and/or introduction should clearly state the claims made, including the contributions made in the paper and important assumptions and limitations. A \answerNo{} or \answerNA{} answer to this question will not be perceived well by the reviewers. 
        \item The claims made should match theoretical and experimental results, and reflect how much the results can be expected to generalize to other settings. 
        \item It is fine to include aspirational goals as motivation as long as it is clear that these goals are not attained by the paper. 
    \end{itemize}

\item {\bf Limitations}
    \item[] Question: Does the paper discuss the limitations of the work performed by the authors?
    \item[] Answer: \answerYes{}
    \item[] Justification: The paper includes a dedicated Limitations section discussing token-level spatial granularity, the binary static--dynamic formulation, lack of fine-grained motion quantities such as displacement or velocity, and semantic leakage failure cases. Additional qualitative examples are provided in Appendix~\ref{app:qualitative_results}, while SAM2 refinement dependence is quantified separately in Tables~\ref{tab:sam2_refinement_mini} and~\ref{tab:additional_pmd_backbones_compact}.
    \item[] Guidelines:
    \begin{itemize}
        \item The answer \answerNA{} means that the paper has no limitation while the answer \answerNo{} means that the paper has limitations, but those are not discussed in the paper. 
        \item The authors are encouraged to create a separate ``Limitations'' section in their paper.
        \item The paper should point out any strong assumptions and how robust the results are to violations of these assumptions (e.g., independence assumptions, noiseless settings, model well-specification, asymptotic approximations only holding locally). The authors should reflect on how these assumptions might be violated in practice and what the implications would be.
        \item The authors should reflect on the scope of the claims made, e.g., if the approach was only tested on a few datasets or with a few runs. In general, empirical results often depend on implicit assumptions, which should be articulated.
        \item The authors should reflect on the factors that influence the performance of the approach. For example, a facial recognition algorithm may perform poorly when image resolution is low or images are taken in low lighting. Or a speech-to-text system might not be used reliably to provide closed captions for online lectures because it fails to handle technical jargon.
        \item The authors should discuss the computational efficiency of the proposed algorithms and how they scale with dataset size.
        \item If applicable, the authors should discuss possible limitations of their approach to address problems of privacy and fairness.
        \item While the authors might fear that complete honesty about limitations might be used by reviewers as grounds for rejection, a worse outcome might be that reviewers discover limitations that aren't acknowledged in the paper. The authors should use their best judgment and recognize that individual actions in favor of transparency play an important role in developing norms that preserve the integrity of the community. Reviewers will be specifically instructed to not penalize honesty concerning limitations.
    \end{itemize}

\item {\bf Theory assumptions and proofs}
    \item[] Question: For each theoretical result, does the paper provide the full set of assumptions and a complete (and correct) proof?
    \item[] Answer: \answerNA{}
    \item[] Justification: The paper does not present formal theoretical results or theorems. Appendix~\ref{app:capacity_analysis} provides a capacity-based explanatory analysis using standard learning-theoretic scaling arguments, but it is not used as a formal theorem or proof.
    \item[] Guidelines:
    \begin{itemize}
        \item The answer \answerNA{} means that the paper does not include theoretical results. 
        \item All the theorems, formulas, and proofs in the paper should be numbered and cross-referenced.
        \item All assumptions should be clearly stated or referenced in the statement of any theorems.
        \item The proofs can either appear in the main paper or the supplemental material, but if they appear in the supplemental material, the authors are encouraged to provide a short proof sketch to provide intuition. 
        \item Inversely, any informal proof provided in the core of the paper should be complemented by formal proofs provided in appendix or supplemental material.
        \item Theorems and Lemmas that the proof relies upon should be properly referenced. 
    \end{itemize}

    \item {\bf Experimental result reproducibility}
    \item[] Question: Does the paper fully disclose all the information needed to reproduce the main experimental results of the paper to the extent that it affects the main claims and/or conclusions of the paper (regardless of whether the code and data are provided or not)?
    \item[] Answer: \answerYes{}
    \item[] Justification: The paper describes the probe formulation, PMD inference procedure, selected blocks, training and validation data, hyperparameters, mask extraction, SAM2 prompt construction, and evaluation protocols. Further details are provided in Section~\ref{sec:setup} and Appendix~\ref{app:implementation}.
    \item[] Guidelines:
    \begin{itemize}
        \item The answer \answerNA{} means that the paper does not include experiments.
        \item If the paper includes experiments, a \answerNo{} answer to this question will not be perceived well by the reviewers: Making the paper reproducible is important, regardless of whether the code and data are provided or not.
        \item If the contribution is a dataset and\slash or model, the authors should describe the steps taken to make their results reproducible or verifiable. 
        \item Depending on the contribution, reproducibility can be accomplished in various ways. For example, if the contribution is a novel architecture, describing the architecture fully might suffice, or if the contribution is a specific model and empirical evaluation, it may be necessary to either make it possible for others to replicate the model with the same dataset, or provide access to the model. In general. releasing code and data is often one good way to accomplish this, but reproducibility can also be provided via detailed instructions for how to replicate the results, access to a hosted model (e.g., in the case of a large language model), releasing of a model checkpoint, or other means that are appropriate to the research performed.
        \item While NeurIPS does not require releasing code, the conference does require all submissions to provide some reasonable avenue for reproducibility, which may depend on the nature of the contribution. For example
        \begin{enumerate}
            \item If the contribution is primarily a new algorithm, the paper should make it clear how to reproduce that algorithm.
            \item If the contribution is primarily a new model architecture, the paper should describe the architecture clearly and fully.
            \item If the contribution is a new model (e.g., a large language model), then there should either be a way to access this model for reproducing the results or a way to reproduce the model (e.g., with an open-source dataset or instructions for how to construct the dataset).
            \item We recognize that reproducibility may be tricky in some cases, in which case authors are welcome to describe the particular way they provide for reproducibility. In the case of closed-source models, it may be that access to the model is limited in some way (e.g., to registered users), but it should be possible for other researchers to have some path to reproducing or verifying the results.
        \end{enumerate}
    \end{itemize}

    \item {\bf Open access to data and code}
    \item[] Question: Does the paper provide open access to the data and code, with sufficient instructions to faithfully reproduce the main experimental results, as described in supplemental material?
    \item[] Answer: \answerYes{}
    \item[] Justification: We provide anonymized code and detailed reproduction instructions in the supplementary material. The supplementary package explains how to obtain the original public datasets from official sources, includes the sampled split files for the YouTube-VOS subsets used in our experiments, and provides scripts or exact commands for reproducing the main training, inference, and evaluation pipelines covered by the release. For results that rely on additional downloaded artifacts or pretrained checkpoints, we explicitly document the required files, expected directory layout, and reproduction steps.
    \item[] Guidelines:
    \begin{itemize}
        \item The answer \answerNA{} means that paper does not include experiments requiring code.
        \item Please see the NeurIPS code and data submission guidelines (\url{https://neurips.cc/public/guides/CodeSubmissionPolicy}) for more details.
        \item While we encourage the release of code and data, we understand that this might not be possible, so \answerNo{} is an acceptable answer. Papers cannot be rejected simply for not including code, unless this is central to the contribution (e.g., for a new open-source benchmark).
        \item The instructions should contain the exact command and environment needed to run to reproduce the results. See the NeurIPS code and data submission guidelines (\url{https://neurips.cc/public/guides/CodeSubmissionPolicy}) for more details.
        \item The authors should provide instructions on data access and preparation, including how to access the raw data, preprocessed data, intermediate data, and generated data, etc.
        \item The authors should provide scripts to reproduce all experimental results for the new proposed method and baselines. If only a subset of experiments are reproducible, they should state which ones are omitted from the script and why.
        \item At submission time, to preserve anonymity, the authors should release anonymized versions (if applicable).
        \item Providing as much information as possible in supplemental material (appended to the paper) is recommended, but including URLs to data and code is permitted.
    \end{itemize}

    \item {\bf Experimental setting/details}
    \item[] Question: Does the paper specify all the training and test details (e.g., data splits, hyperparameters, how they were chosen, type of optimizer) necessary to understand the results?
    \item[] Answer: \answerYes{}
    \item[] Justification: The paper specifies the training and validation splits, evaluation benchmarks, optimizer, learning rate, number of epochs, sampled token pairs, selected blocks, mask thresholding, SAM2 prompt construction, inference context, and evaluation metrics. These details are provided in Section~\ref{sec:setup} and Appendix~\ref{app:implementation}.
    \item[] Guidelines:
    \begin{itemize}
        \item The answer \answerNA{} means that the paper does not include experiments.
        \item The experimental setting should be presented in the core of the paper to a level of detail that is necessary to appreciate the results and make sense of them.
        \item The full details can be provided either with the code, in appendix, or as supplemental material.
    \end{itemize}

\item {\bf Experiment statistical significance}
    \item[] Question: Does the paper report error bars suitably and correctly defined or other appropriate information about the statistical significance of the experiments?
    \item[] Answer: \answerNo{}
    \item[] Justification: The main benchmark tables do not report confidence intervals, statistical tests, or error bars for all methods and backbones. Appendix~\ref{app:additional_quantitative} reports mean and one-standard-deviation bands for the data-starvation experiment when multiple subset seeds are available, but this uncertainty analysis does not cover all main benchmark results.
    \item[] Guidelines:
    \begin{itemize}
        \item The answer \answerNA{} means that the paper does not include experiments.
        \item The authors should answer \answerYes{} if the results are accompanied by error bars, confidence intervals, or statistical significance tests, at least for the experiments that support the main claims of the paper.
        \item The factors of variability that the error bars are capturing should be clearly stated (for example, train/test split, initialization, random drawing of some parameter, or overall run with given experimental conditions).
        \item The method for calculating the error bars should be explained (closed form formula, call to a library function, bootstrap, etc.)
        \item The assumptions made should be given (e.g., Normally distributed errors).
        \item It should be clear whether the error bar is the standard deviation or the standard error of the mean.
        \item It is OK to report 1-sigma error bars, but one should state it. The authors should preferably report a 2-sigma error bar than state that they have a 96\% CI, if the hypothesis of Normality of errors is not verified.
        \item For asymmetric distributions, the authors should be careful not to show in tables or figures symmetric error bars that would yield results that are out of range (e.g., negative error rates).
        \item If error bars are reported in tables or plots, the authors should explain in the text how they were calculated and reference the corresponding figures or tables in the text.
    \end{itemize}

\item {\bf Experiments compute resources}
    \item[] Question: For each experiment, does the paper provide sufficient information on the computer resources (type of compute workers, memory, time of execution) needed to reproduce the experiments?
    \item[] Answer: \answerYes{}
    \item[] Justification: The paper reports the hardware setup, including GPU, CPU, memory configuration, trainable parameter counts, and inference FLOP overhead in Section~\ref{sec:setup} and Appendices~\ref{app:implementation} and~\ref{app:efficiency}. All experiments are conducted on a single consumer-grade GPU workstation without distributed training.
    \item[] Guidelines:
    \begin{itemize}
        \item The answer \answerNA{} means that the paper does not include experiments.
        \item The paper should indicate the type of compute workers CPU or GPU, internal cluster, or cloud provider, including relevant memory and storage.
        \item The paper should provide the amount of compute required for each of the individual experimental runs as well as estimate the total compute. 
        \item The paper should disclose whether the full research project required more compute than the experiments reported in the paper (e.g., preliminary or failed experiments that didn't make it into the paper). 
    \end{itemize}
    
\item {\bf Code of ethics}
    \item[] Question: Does the research conducted in the paper conform, in every respect, with the NeurIPS Code of Ethics \url{https://neurips.cc/public/EthicsGuidelines}?
    \item[] Answer: \answerYes{}
    \item[] Justification: The research uses public datasets and pretrained models, does not involve private data collection or human-subject experiments, and discusses potential misuse risks. Broader impacts and potential negative uses are addressed in Appendix~\ref{app:broader_impacts}.
    \item[] Guidelines:
    \begin{itemize}
        \item The answer \answerNA{} means that the authors have not reviewed the NeurIPS Code of Ethics.
        \item If the authors answer \answerNo, they should explain the special circumstances that require a deviation from the Code of Ethics.
        \item The authors should make sure to preserve anonymity (e.g., if there is a special consideration due to laws or regulations in their jurisdiction).
    \end{itemize}

\item {\bf Broader impacts}
    \item[] Question: Does the paper discuss both potential positive societal impacts and negative societal impacts of the work performed?
    \item[] Answer: \answerYes{}
    \item[] Justification: Appendix~\ref{app:broader_impacts} discusses positive impacts such as improving accessibility and computational efficiency for dynamic scene understanding, as well as potential negative impacts such as misuse for unauthorized surveillance.
    \item[] Guidelines:
    \begin{itemize}
        \item The answer \answerNA{} means that there is no societal impact of the work performed.
        \item If the authors answer \answerNA{} or \answerNo, they should explain why their work has no societal impact or why the paper does not address societal impact.
        \item Examples of negative societal impacts include potential malicious or unintended uses (e.g., disinformation, generating fake profiles, surveillance), fairness considerations (e.g., deployment of technologies that could make decisions that unfairly impact specific groups), privacy considerations, and security considerations.
        \item The conference expects that many papers will be foundational research and not tied to particular applications, let alone deployments. However, if there is a direct path to any negative applications, the authors should point it out. For example, it is legitimate to point out that an improvement in the quality of generative models could be used to generate Deepfakes for disinformation. On the other hand, it is not needed to point out that a generic algorithm for optimizing neural networks could enable people to train models that generate Deepfakes faster.
        \item The authors should consider possible harms that could arise when the technology is being used as intended and functioning correctly, harms that could arise when the technology is being used as intended but gives incorrect results, and harms following from (intentional or unintentional) misuse of the technology.
        \item If there are negative societal impacts, the authors could also discuss possible mitigation strategies (e.g., gated release of models, providing defenses in addition to attacks, mechanisms for monitoring misuse, mechanisms to monitor how a system learns from feedback over time, improving the efficiency and accessibility of ML).
    \end{itemize}
    
\item {\bf Safeguards}
    \item[] Question: Does the paper describe safeguards that have been put in place for responsible release of data or models that have a high risk for misuse (e.g., pre-trained language models, image generators, or scraped datasets)?
    \item[] Answer: \answerNA{}
    \item[] Justification: The paper does not release a new high-risk pretrained model, image generator, scraped dataset, or other asset requiring controlled-release safeguards. The work uses public datasets and existing pretrained models, and potential misuse of motion segmentation is discussed in Appendix~\ref{app:broader_impacts}.
    \item[] Guidelines:
    \begin{itemize}
        \item The answer \answerNA{} means that the paper poses no such risks.
        \item Released models that have a high risk for misuse or dual-use should be released with necessary safeguards to allow for controlled use of the model, for example by requiring that users adhere to usage guidelines or restrictions to access the model or implementing safety filters. 
        \item Datasets that have been scraped from the Internet could pose safety risks. The authors should describe how they avoided releasing unsafe images.
        \item We recognize that providing effective safeguards is challenging, and many papers do not require this, but we encourage authors to take this into account and make a best faith effort.
    \end{itemize}

\item {\bf Licenses for existing assets}
    \item[] Question: Are the creators or original owners of assets (e.g., code, data, models), used in the paper, properly credited and are the license and terms of use explicitly mentioned and properly respected?
    \item[] Answer: \answerYes{}
    \item[] Justification: The paper credits the existing datasets, pretrained models, and tools through citations, including YouTube-VOS, DAVIS, SegTrackv2, FBMS-59, SAM2, and feed-forward 3D foundation models. Appendix~\ref{app:asset_licenses} lists these assets and their license or terms-of-use information where publicly available; for SegTrackv2, whose official page does not explicitly state a standard license, we state this limitation and use the dataset only for research evaluation.
    \item[] Guidelines:
    \begin{itemize}
        \item The answer \answerNA{} means that the paper does not use existing assets.
        \item The authors should cite the original paper that produced the code package or dataset.
        \item The authors should state which version of the asset is used and, if possible, include a URL.
        \item The name of the license (e.g., CC-BY 4.0) should be included for each asset.
        \item For scraped data from a particular source (e.g., website), the copyright and terms of service of that source should be provided.
        \item If assets are released, the license, copyright information, and terms of use in the package should be provided. For popular datasets, \url{paperswithcode.com/datasets} has curated licenses for some datasets. Their licensing guide can help determine the license of a dataset.
        \item For existing datasets that are re-packaged, both the original license and the license of the derived asset (if it has changed) should be provided.
        \item If this information is not available online, the authors are encouraged to reach out to the asset's creators.
    \end{itemize}

\item {\bf New assets}
    \item[] Question: Are new assets introduced in the paper well documented and is the documentation provided alongside the assets?
    \item[] Answer: \answerYes{}
    \item[] Justification: The supplementary material includes anonymized code and sampled split files used for reproduction. These assets are documented in the accompanying README files, which describe the expected directory structure, required external dependencies, dataset preparation, and the commands needed to reproduce the released experiments. We do not release a new dataset or benchmark; the released assets are intended to support reproducibility of the method and experiments in the paper.
    \item[] Guidelines:
    \begin{itemize}
        \item The answer \answerNA{} means that the paper does not release new assets.
        \item Researchers should communicate the details of the dataset\slash code\slash model as part of their submissions via structured templates. This includes details about training, license, limitations, etc. 
        \item The paper should discuss whether and how consent was obtained from people whose asset is used.
        \item At submission time, remember to anonymize your assets (if applicable). You can either create an anonymized URL or include an anonymized zip file.
    \end{itemize}

\item {\bf Crowdsourcing and research with human subjects}
    \item[] Question: For crowdsourcing experiments and research with human subjects, does the paper include the full text of instructions given to participants and screenshots, if applicable, as well as details about compensation (if any)? 
    \item[] Answer: \answerNA{}
    \item[] Justification: The paper does not involve crowdsourcing experiments or research with human subjects.
    \item[] Guidelines:
    \begin{itemize}
        \item The answer \answerNA{} means that the paper does not involve crowdsourcing nor research with human subjects.
        \item Including this information in the supplemental material is fine, but if the main contribution of the paper involves human subjects, then as much detail as possible should be included in the main paper. 
        \item According to the NeurIPS Code of Ethics, workers involved in data collection, curation, or other labor should be paid at least the minimum wage in the country of the data collector. 
    \end{itemize}

\item {\bf Institutional review board (IRB) approvals or equivalent for research with human subjects}
    \item[] Question: Does the paper describe potential risks incurred by study participants, whether such risks were disclosed to the subjects, and whether Institutional Review Board (IRB) approvals (or an equivalent approval/review based on the requirements of your country or institution) were obtained?
    \item[] Answer: \answerNA{}
    \item[] Justification: The paper does not involve crowdsourcing or human-subject research, so IRB approval is not applicable.
    \item[] Guidelines:
    \begin{itemize}
        \item The answer \answerNA{} means that the paper does not involve crowdsourcing nor research with human subjects.
        \item Depending on the country in which research is conducted, IRB approval (or equivalent) may be required for any human subjects research. If you obtained IRB approval, you should clearly state this in the paper. 
        \item We recognize that the procedures for this may vary significantly between institutions and locations, and we expect authors to adhere to the NeurIPS Code of Ethics and the guidelines for their institution. 
        \item For initial submissions, do not include any information that would break anonymity (if applicable), such as the institution conducting the review.
    \end{itemize}

\item {\bf Declaration of LLM usage}
    \item[] Question: Does the paper describe the usage of LLMs if it is an important, original, or non-standard component of the core methods in this research? Note that if the LLM is used only for writing, editing, or formatting purposes and does \emph{not} impact the core methodology, scientific rigor, or originality of the research, declaration is not required.
    \item[] Answer: \answerNA{}
    \item[] Justification: LLMs were used only for writing refinement, language editing, formatting assistance, and non-core utility code such as visualization or file-I/O helpers. LLMs were not used as part of the core method, experimental design, data generation, analysis, or scientific claims; therefore, declaration is not required under the checklist guidance.
    \item[] Guidelines:
    \begin{itemize}
        \item The answer \answerNA{} means that the core method development in this research does not involve LLMs as any important, original, or non-standard components.
        \item Please refer to our LLM policy in the NeurIPS handbook for what should or should not be described.
    \end{itemize}

\end{enumerate}

\end{document}